\documentclass{article}
\usepackage[preprint]{colm2025_conference}
\usepackage{multicol}

\usepackage{microtype}
\usepackage{makecell}
\usepackage{subcaption} 
\usepackage{longtable}
\usepackage[T1]{fontenc}    
\definecolor{green}{RGB}{0,150,10}
\definecolor{blue}{RGB}{0,148,181}
\definecolor{orange}{RGB}{194,153,107}
\usepackage[colorlinks,linkcolor=red,anchorcolor=green,citecolor=blue]{hyperref}
\usepackage{url}            
\usepackage{lmodern}
\usepackage{fontawesome}

\usepackage{lineno}
\usepackage{tabularx}

\usepackage{booktabs}       
\usepackage{amsfonts}       
\usepackage{nicefrac}       
\usepackage{xcolor}         
\definecolor{background-grey}{RGB}{220,220,220}  
\definecolor{cell-green}{RGB}{221, 255, 225}  
\definecolor{cell-red}{RGB}{255, 224, 224}  
\definecolor{light-green}{HTML}{A2D9A2}
\definecolor{llight-green}{HTML}{C7EFCF}
\definecolor{light-red}{HTML}{FFD1D1}
\definecolor{light-orange}{HTML}{FFC9A3}

\usepackage{amsmath}
\usepackage{marvosym}
\usepackage{graphicx}
\usepackage{colortbl}
\usepackage{multirow}
\usepackage{subcaption}
\usepackage{changes}
\usepackage{bookmark}
\usepackage{listings}
\usepackage{wrapfig}
\lstdefinelanguage{Dialogue}{
  morekeywords={Influencer,Voter,rating},
  sensitive=false,
  morecomment=[l]{//},
}
\usepackage{natbib}
\usepackage{bibentry}     
\nobibliography*          

\usepackage{mdframed}
\usepackage{caption}
\usepackage{fancyhdr}
\usepackage{datetime}
\usepackage{adjustbox}
\usepackage{amssymb}
\usepackage{tcolorbox}
\usepackage{makecell}
\usepackage{multirow}      
\usepackage{array}         
\usepackage{fancyvrb}
\usepackage{fvextra}
\usepackage{pifont}
\usepackage{tikz}
\usepackage{wasysym}
\usepackage{placeins}

\usepackage{geometry}
\newcommand{\name}{DeepSight}

\definecolor{Blue4Head}{RGB}{58,104,153}

\title{\centering \bf \Large \name: An All-in-One LM Safety Toolkit}

\author{
\scalebox{0.84}{
\quad \quad \quad \quad \quad \quad \quad \quad
Bo Zhang$^{*}$,
Jiaxuan Guo$^{*}$,
Lijun Li$^{*}$,
Dongrui Liu$^{*}$,
Sujin Chen, 
Guanxu Chen,
}\\
\scalebox{0.84}{
\textbf{
 \quad \quad \quad \quad \quad \quad
Zhijie Zheng,
Qihao Lin, 
Lewen Yan, 
Chen Qian, 
Yijin Zhou, 
Yuyao Wu,
Shaoxiong Guo, 
}
}\\
\scalebox{0.84}{
\textbf{
\quad \quad \quad \quad \quad \quad \quad \quad
Tianyi Du,
Jingyi Yang, 
Xuhao Hu, 
Ziqi Miao, 
Xiaoya Lu,
Jing Shao\textsuperscript{\Letter},
Xia Hu}
}
\vspace{20pt}\\
\scalebox{0.9}{\quad \quad \quad \quad \quad \quad  \quad \quad \quad \quad \quad \quad \quad \quad \quad \quad \quad \quad \large \textbf{Shanghai AI Laboratory} }
}

\renewcommand{\thefootnote}{\fnsymbol{footnote}}
\newcommand\blfootnote[1]{%
\begingroup
\renewcommand\thefootnote{}\footnote{#1}%
\addtocounter{footnote}{-1}%
\endgroup
}

\begin{document}

\hypersetup{
    linkcolor=black,
    filecolor=black,      
    urlcolor=blue,
    citecolor=blue,
}

\maketitle
\blfootnote{
* Co-leads; \Letter\  Corresponding author. 
}

\begin{center}
    \vspace{-26pt}
    \small
    \href{https://github.com/AI45Lab/DeepSafe}{\textcolor{black!55}{\faGithub}\ \texttt{github.com/AI45Lab/DeepSafe}}
    \quad$\vert$\quad
    \href{https://github.com/AI45Lab/DeepScan/}{\textcolor{black!55}{\faGithub}\ \texttt{github.com/AI45Lab/DeepScan}}
    \vspace{26pt}
\end{center}

\begin{abstract}
As the development of Large Models (LMs) progresses rapidly, their safety is also a priority. In current Large Language Models (LLMs) and Multimodal Large Language Models (MLLMs) safety workflow, evaluation, diagnosis, and alignment are often handled by separate tools. Specifically, safety evaluation can only locate external behavioral risks but cannot figure out internal root causes. Meanwhile, safety diagnosis often drifts from concrete risk scenarios and remains at the explainable level. In this way, safety alignment lack dedicated explanations of changes in internal mechanisms, potentially degrading general capabilities. To systematically address these issues, we propose an open-source project, namely \textbf{DeepSight}, to practice a new safety evaluation-diagnosis integrated paradigm. DeepSight is low-cost, reproducible, efficient, and highly scalable large-scale model safety evaluation project consisting of a evaluation toolkit DeepSafe and a diagnosis toolkit DeepScan. By unifying task and data protocols, we build a connection between the two stages and transform safety evaluation from black-box to white-box insight. Besides, DeepSight is the first open source toolkit that support the frontier AI risk evaluation and joint safety evaluation and diagnosis.

\end{abstract}

\newpage

\tableofcontents

\newpage

\hypersetup{
    linkcolor=red,
    filecolor=black,      
    urlcolor=blue,
    citecolor=blue,
}

\setcounter{section}{0}

\section{Introduction}\label{sec:intro}
The rapid evolution of Large Language Models (LLMs) and Multimodal Large Language Models (MLLMs) has fundamentally transformed the artificial intelligence landscape, positioning safety as a paramount concern alongside performance capability~\citep{jailbroken}. As these models are increasingly integrated into critical applications~\citep{Llm_for_education,llm_psychology,yang2026toward}, the industry faces an urgent need to ensure they operate within secure and ethical boundaries.

However, the current landscape of safety evaluation remains fractured between black-box assessment and white-box insight. While the industry has established robust evaluation frameworks, ranging from OpenAI Evals~\citep{openai_evals} and Inspect~\citep{inspect_ai} to comprehensive platforms like OpenCompass~\citep{opencompass2023} and HELM~\citep{liang2022helm}, these tools primarily focus on quantifying behavior capabilities instead of safety mechanisms. Conversely, diagnostic research has made significant strides in decoding internal mechanisms, including probing latent geometric boundaries, identifying safety-specific neurons, and analyzing reasoning dynamics via information flow~\citep{spin,mipeaks,burns2022discovering, zou2023representation}. Yet, these diagnostic methods often operate in isolation from standardized benchmarks, treating internal representation analysis as a separate academic pursuit rather than a debugging tool for deployment risks.

To systematically address these issues, we propose \textbf{DeepSight}, an open-source project, that proposes and implements a new safety evaluation-diagnosis integrated paradigm. By unifying task and data protocols, DeepSight connects evaluation and diagnosis. In this way, researchers and developers not only identify what went wrong but also understand how safety concepts are encoded internally, facilitating more reliable and interpretable repairs. DeepSight is composed of two core engines: 
\begin{itemize}
    \item \textbf{DeepSafe}: A modular, configuration-driven framework for multimodal LLM/MLLM evaluation that integrates over 20 safety benchmarks. More importantly, DeepSafe also supports the evaluation of frontier AI risks, \emph{i.e.,} high-severity risks~\citep{SafeWork-F1}. It provides a low-cost, reproducible, efficient, and highly scalable large-scale model security evaluation toolkit for researchers and professional developers.
    \item \textbf{DeepScan}: A standardized and extensible framework equipped with a suite of LLM diagnostic tools. It efficiently locates key layers and neurons to probe representation-level structures and objective-level conflicts without modifying model weights.
\end{itemize}

To the best of our knowledge, DeepSight is the \textbf{first open source toolkit} that support the \textbf{frontier AI risk evaluation} and \textbf{joint safety evaluation and diagnosis}. We conduct a comprehensive analysis of prevailing Large Models (LMs), revealing safety trends in the current LM landscape:
\begin{itemize}
    \item The introduction of visual modalities significantly expands the attack surface, causing safety alignment to decline across all model tiers compared to text-only scenarios. Furthermore, this transition amplifies the safety performance disparity between open-source and closed-source models, with the latter maintaining a distinct advantage in cross-modal scenarios.
    \item We observe a complex trade-off regarding reasoning capabilities. In multimodal environments, reasoning-enabled models demonstrate superior safety alignment, effectively identifying image-text splitting attacks where non-reasoning models fail.
    \item For AI frontier risks, safety advantages prove non-transferable across dimensions. No single model dominates every category, and even top-ranking models can exhibit severe failures in specific risks, for example Kimi-K2-Thinking ranks last in Manipulation, while lower-tier models can surprisingly lead in some dimensions.
    \item Diagnosis analysis reveals that effective safety depends on the precise geometric structure of a model's latent space, where both insufficient and excessive separation between safe and harmful representations prove detrimental to model robustness.
\end{itemize}

\section{DeepSight Framework}\label{sec:framework}

\subsection{Overview}\label{subsec:pre}
While traditional evaluation often isolates evaluation and diagnosis, resulting in misaligned goals where evaluation identifies external risks without addressing internal root causes, DeepSight connects them into a verifiable engineering loop, effectively transitioning from black-box assessment to white-box insight. The framework is composed of two engines: DeepSafe, a configuration-driven framework that standardizes behavioral assessment across over 20 benchmarks, and DeepScan, a diagnostic tool that probes representation-level structures and objective-level conflicts to explain the mechanisms behind observed failures. By harmonizing task and data protocols, DeepSight enables researchers not only to identify what went wrong but also to understand how safety concepts are encoded internally, facilitating more reliable and interpretable repairs.

\subsection{DeepSafe}
\begin{figure}[h]
    \centering
    \includegraphics[width=\linewidth]{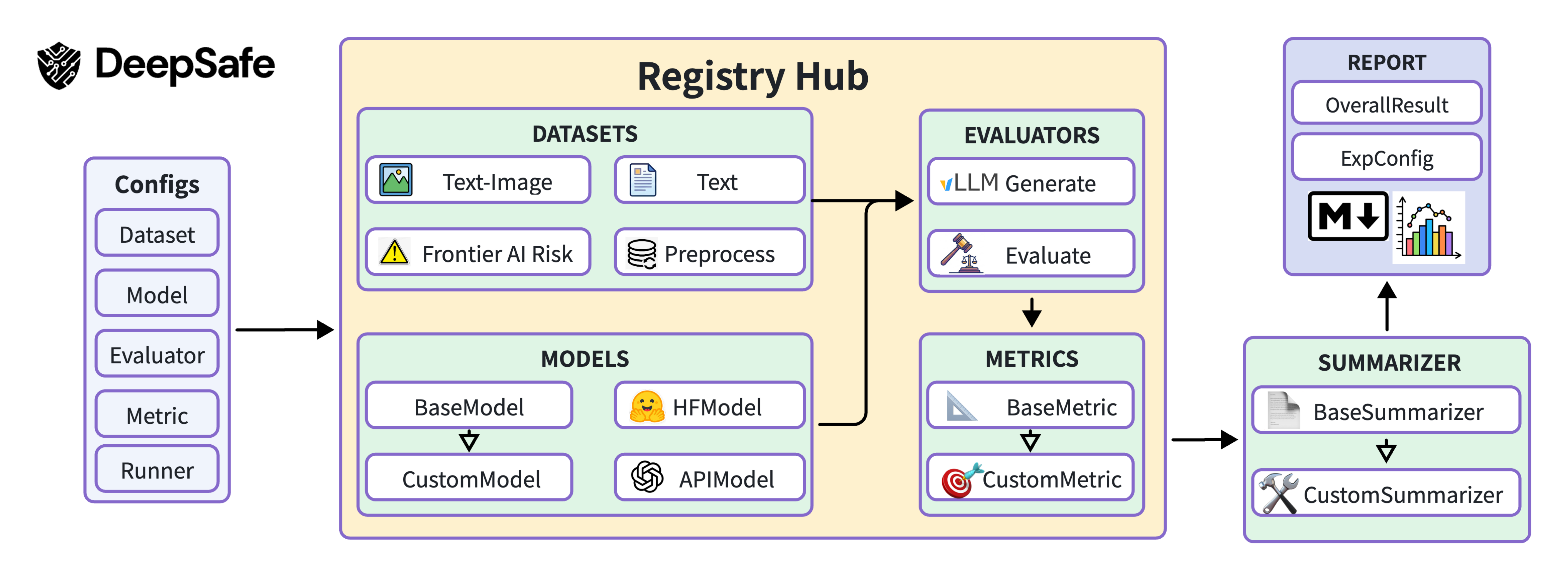}
    \caption{\textbf{The DeepSafe Architecture.} DeepSafe employs a configuration-driven approach where the \textit{Registry Hub} orchestrates the interaction between Datasets, Models, and Evaluators. This modular design automates the workflow from inference (Runner) to analysis (Summarizer), producing standardized reports.}
    \label{fig:deepsafe_arch}
\end{figure}
Safety evaluation for large models currently suffers from a lack of standardized protocols and dedicated assessment tools. We present \textbf{DeepSafe}, an all-in-one framework integrating over 20 safety datasets, such as SALAD-Bench~\citep{salad} and HarmBench~\citep{harmbench}, alongside the specialized ProGuard judge model~\citep{proguard}. DeepSafe supports LLM/MLLM evaluation through a modular, configuration-driven architecture (illustrated in Figure~\ref{fig:deepsafe_arch}) that automates the entire workflow from inference to deep reporting. It is designed to advance LM safety evaluation from outcome-based testing to rigorous analysis, accelerating the development of trustworthy AI evaluation. The core components of the framework are summarized in Table~\ref{tab:components}.

\begin{table}[h]
\centering
\renewcommand{\arraystretch}{1.2}
\small 
\resizebox{0.88\linewidth}{!}{
\begin{tabular}{l p{5cm} p{6cm}}
\hline
\textbf{Component} & \textbf{Role} & \textbf{Key Implementations} \\
\hline
\textbf{Models} & 
Unified interface for model loading, management, and high-performance inference. & 
\texttt{HFModel} (Hugging Face), \texttt{APIModel} (e.g., GPT-5), and vLLM integration for accelerated generation. \\

\textbf{Datasets} & 
Standardized data handling supporting various modalities and safety domains. & 
Multimodal support, Frontier AI Risk datasets, and Preprocess pipelines. \\

\textbf{Evaluators} & 
Executes safety assessments compatible with diverse benchmark protocols. & 
Native evaluators, rule-based matching, and model-based judges. \\

\textbf{Runner} & 
Orchestrates the end-to-end workflow from configuration to execution. & 
Configuration-driven architecture and automated pipeline management. \\

\textbf{Summarizer} & 
Aggregates evaluation metrics and generates visualization reports. & 
\texttt{BaseSummarizer} for statistical analysis and automatic Markdown report generation. \\
\hline
\end{tabular}
}
\caption{Summary of DeepSafe's core components. The framework adopts a modular design enabling flexible extension and high-performance execution, supporting native evaluators for various benchmarks.}
\label{tab:components}
\end{table}

As listed in Table~\ref{tab:components}, DeepSafe is built upon five foundational modules, coordinated via a centralized Registry Mechanism. This design decouples task implementation from execution logic, allowing researchers to integrate new datasets or metrics with minimal code changes.

\subsubsection{Core Modules}

\paragraph{Models.}
The Models component provides a unified interface for model inference, abstracting away differences between local weights and remote APIs. We implement \texttt{VLLMLocalModel} based on the vLLM library to enable high-throughput generation for open-source models. For closed-source models, \texttt{APIModel} encapsulates interactions with commercial APIs. This abstraction allows seamless switching between backends by modifying the configuration.

\paragraph{Datasets.}
DeepSafe standardizes data handling through the \texttt{BaseDataset} protocol. Regardless of the raw format, all data loaders normalize inputs into a consistent schema containing unique identifiers, prompts, and reference answers. The framework currently integrates loaders for over 20 benchmarks, covering diverse domains from traditional content safety to frontier AI risks. Additionally, \texttt{JsonlDataset} allows users to swiftly plug in custom datasets.

\paragraph{Runners.}
The Runner serves as the orchestration engine, managing the end-to-end evaluation lifecycle. We provide a robust \texttt{LocalRunner} that handles pipeline automation, including batch processing and result persistence. A key feature is its state-aware execution, which automatically detects and skips generation process, enabling efficient resumption of evaluations.

\paragraph{Evaluators \& Metrics.}
Evaluation logic is decoupled into Evaluators and Metrics. DeepSafe supports a hybrid evaluation paradigm to accommodate diverse benchmarks:
(1) \textbf{Native Evaluators}: We integrate official evaluation scripts for specific benchmarks (e.g., HarmBench, BeaverTails) to ensure strict adherence to their original protocols.
(2) \textbf{Rule-based Evaluators}: For tasks with deterministic criteria, we provide fast keyword matching and regular expression handlers.
(3) \textbf{Model-based Evaluators}: We support LLM-as-a-Judge, leveraging strong models to assess complex responses where rule-based methods fail.
Post-evaluation, Metrics modules aggregate raw judgments into statistical scores, ensuring precise measurement.

\paragraph{Summarizers.}
To facilitate rapid analysis, the Summarizer component aggregates evaluation metrics and renders them into human-readable formats. The \texttt{StandardSummarizer} automatically generates a comprehensive Markdown report containing summary tables and configuration details, along with a structured JSON file for downstream analysis.

\subsubsection{ProGuard: A Specialized Safety Evaluator}\label{subsec:proguard}
While the framework supports various native and general-purpose evaluators, generic LLMs often lack specific alignment for safety nuances. DeepSafe therefore integrates \textbf{ProGuard}~\citep{proguard}, a specialized safety evaluator model designed to deliver superior judgment accuracy. ProGuard is fine-tuned on a curated dataset of 87k safety pairs, enabling it to detect subtle risks and adversarial attacks that generic models might miss. Within the framework, ProGuard functions as a plug-and-play judge backend, offering a high-fidelity alternative to the benchmarks' original evaluators.

\subsubsection{Unified Evaluation Workflow}\label{subsec:workflow}
DeepSafe streamlines the safety evaluation process into a standardized four-stage workflow, fully driven by a single YAML configuration file:

\begin{enumerate}
    \item \textbf{Configuration}: Users define the target model, dataset, and evaluation parameters in a declarative YAML file.
    \item \textbf{Inference}: The Runner initializes the model and dataset, performing batch inference to generate responses.
    \item \textbf{Assessment}: The Evaluator processes the generated responses—using either native protocols, rule-based matching, or ProGuard—to produce judgments.
    \item \textbf{Reporting}: The Summarizer aggregates judgments into final metrics and exports visualization reports.
\end{enumerate}

This ``Config-as-Execution'' paradigm significantly lowers the barrier for safety research, ensuring that evaluations are both scalable and reproducible.

\subsection{DeepScan}

Safety evaluation frameworks such as DeepSafe answer the question of \emph{whether} a model exhibits safe behavior under standardized benchmarks. To support the development of more reliable and interpretable systems, it is equally important to understand \emph{how} safety-related concepts are encoded in internal representations and whether different safety objectives conflict at the parameter level. Such diagnostic analyses require consistent access to intermediate activations, standardized metrics, and reproducible pipelines across model families and benchmarks. We present \textbf{DeepScan}, a flexible and extensible diagnostic framework for LLMs and MLLMs. DeepScan is designed to complement DeepSafe: together they form a complete evaluation--diagnosis engineering pipeline---DeepSafe assesses behavioral safety outcomes, while DeepScan probes representation-level structure and objective-level conflicts without modifying model weights.

\begin{figure}[h]
    \centering
    \includegraphics[width=\linewidth]{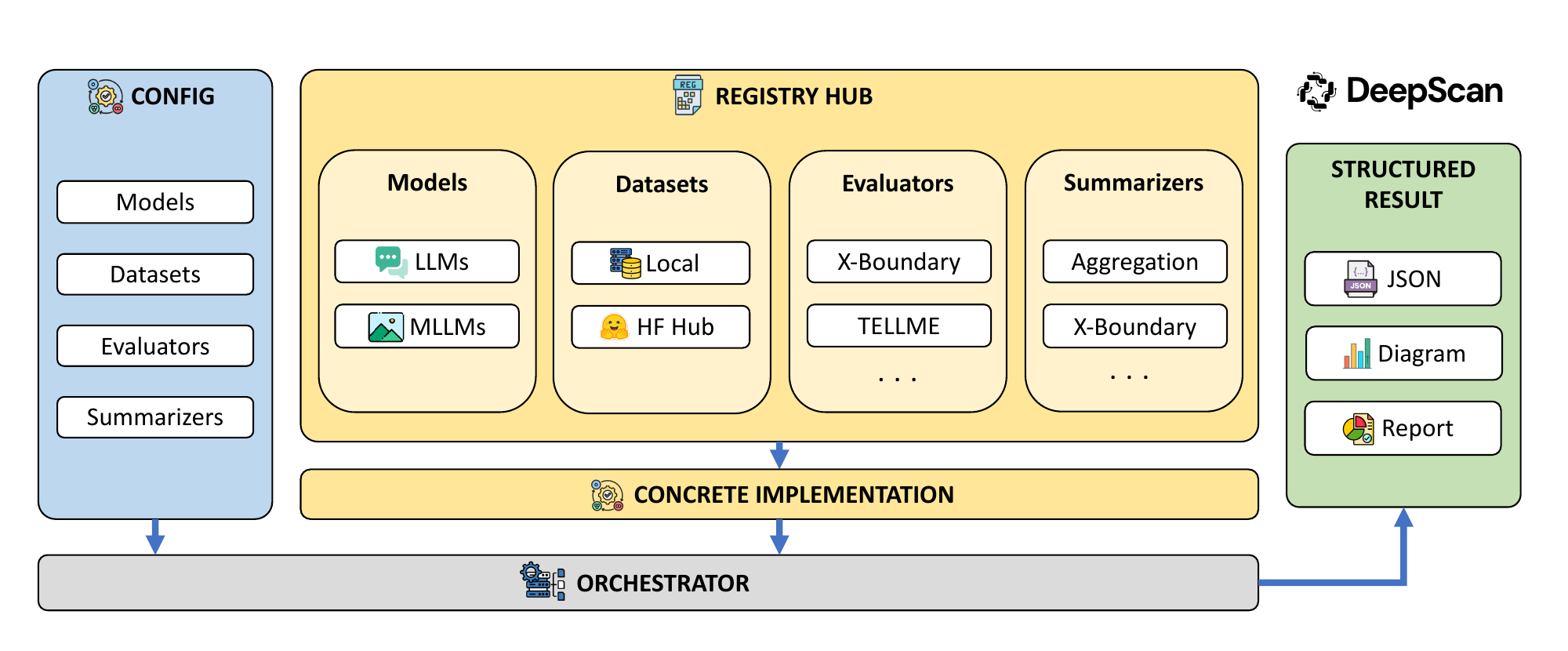}
    \caption{\textbf{The DeepScan Architecture.} DeepScan follows a configuration-driven paradigm, with a Registry Hub coordinating the interplay between Datasets, Models, and Evaluators. This modular organization streamlines the full workflow—from inference execution (Runner/Orchestrator) to aggregation and analytical summarization (Summarizer)—and produces standardized, machine-readable outputs (e.g., JSON) alongside report-oriented artifacts such as diagrams and written summaries.}
    \label{fig:deepsafe_arch}
\end{figure}

DeepScan is built around a ``Register $\rightarrow$ Configure $\rightarrow$ Execute $\rightarrow$ Summarize'' workflow. A centralized registry system decouples model instantiation, dataset loading, evaluator logic, and report generation, so that new model families, diagnostic benchmarks, or metrics can be integrated with minimal changes to core execution code. The entire pipeline is driven by a single declarative configuration file (YAML or JSON), and can be invoked from the command line without writing code, or programmatically via a unified API. The framework exposes a consistent inference interface across supported model families (Qwen, Llama, Mistral, Gemma, GLM, InternLM, InternVL, etc.) while preserving access to the underlying Hugging Face model and tokenizer for representation extraction. Table~\ref{tab:deepscan_components} summarizes the core components; Table~\ref{tab:deepscan_evaluators} lists the built-in diagnostic evaluators and their associated metrics and data sources.

\begin{table}[t]
\centering
\renewcommand{\arraystretch}{1.2}
\small
\begin{tabular}{l p{4.5cm} p{8cm}}
\hline
\textbf{Component} & \textbf{Role} & \textbf{Key Implementations} \\
\hline
\textbf{Registry System} &
Centralized registration and retrieval of models, datasets, evaluators, and summarizers. &
\texttt{BaseRegistry}, \texttt{ModelRegistry}, \texttt{DatasetRegistry}, \texttt{EvaluatorRegistry}, \texttt{SummarizerRegistry}; factory-based instantiation with configurable parameters. \\

\textbf{Configuration} &
Declarative specification of runs via YAML; support for batch runs. &
\texttt{ConfigLoader} from file or dict; dot-notation access; per-evaluator overrides for dataset and summarizer. \\

\textbf{Model Runners} &
Unified inference API and access to raw model/tokenizer for hooks. &
\texttt{BaseModelRunner} with \texttt{generate()} and chat-style APIs; \texttt{model}/\texttt{tokenizer} exposed for activation extraction. \\

\textbf{Evaluators} &
Encapsulation of diagnostic protocols with a single \texttt{evaluate(model, dataset, ...)} contract. &
\texttt{BaseEvaluator}; built-in TELLME \citep{tellme}, X-Boundary \citep{xboundary}, MI-Peaks \citep{mipeaks}, SPIN \citep{spin};. \\

\textbf{Summarizers} &
Aggregation of raw results into structured metrics and human-readable reports. &
\texttt{BaseSummarizer}; benchmark-specific summarizers (e.g., TELLME, X-Boundary, SPIN); combined run-level summarizer. \\
\hline
\end{tabular}
\caption{Summary of DeepScan's core components. The registry-based design allows new models, datasets, and diagnostic methods to be added without altering the execution engine.}
\label{tab:deepscan_components}
\end{table}

\begin{table}[t]
\centering
\renewcommand{\arraystretch}{1.35}
\small
\begin{tabular}{@{}p{2cm}p{5cm}p{8cm}@{}}
\hline
\textbf{Evaluator} & \textbf{Primary metrics} & \textbf{Data \& methodology} \\
\hline
X-Boundary &
separation-score ($\uparrow$); boundary-ratio ($\downarrow$); dist-bound-safe ($\downarrow$); dist-bound-harmful ($\uparrow$) &
Representation diagnostic tool for open-source LLMs focusing on the geometric structure of hidden spaces formed by safe, harmful, and boundary samples. Centroid-based metrics quantify separability and boundary alignment; optional t-SNE per layer. \\
\hline
TELLME &
$r_{\mathrm{diff}}$ ($\uparrow$); $r_{\mathrm{same}}$ ($\downarrow$); eRank; cos\_sim ($\downarrow$); L1, L2, Hausdorff ($\uparrow$) &
Enhances intrinsic transparency by disentangling representations across behaviors. Coding-rate and distance metrics over class-conditioned activations (e.g., filtered BeaverTails); applicable to text/multimodal and Dense/MoE architectures. \\
\hline
SPIN &
Fairness--Privacy Neurons Coupling Index ($\downarrow$) &
Neuron-level diagnosis for open-source LLMs; quantifies overlap between fairness- and privacy-related neurons. Generalizable to other objective pairs (e.g., safety--utility, alignment--generalization). \\
\hline
MI-Peaks &
Mutual Information Trajectory (visualization) &
Information-theoretic framework to examine reasoning dynamics during generation. Tracks MI between step-wise hidden representations and the correct answer; MI Peaks as structural signature. For Transformer-based autoregressive LLMs/LRMs. \\
\hline
\end{tabular}
\caption{Built-in DeepScan diagnostic evaluators (metric names and directions calibrated). Each implements \texttt{BaseEvaluator} and can be combined in a single run with different datasets.}
\label{tab:deepscan_evaluators}
\end{table}
\subsubsection{Core Components}\label{subsec:deepscan_components}

\paragraph{Registry system.}
DeepScan adopts a registry pattern for all pluggable components. The \textbf{Model Registry} stores model factories keyed by family and generation (e.g., \texttt{qwen3} \citep{qwen3}, \texttt{llama3} \citep{llama3}); lookup returns a \textbf{model runner} that provides a uniform \texttt{generate()} interface and retains references to the underlying Hugging Face model and tokenizer. The \textbf{Dataset Registry} maps dataset names (e.g., \texttt{tellme/beaver\_tails\_filtered}, \texttt{xboundary/diagnostic}) to loaders that return a consistent structure (e.g., splits keyed by \texttt{train}/\texttt{test}, or message lists for chat templates). The \textbf{Evaluator Registry} and \textbf{Summarizer Registry} map string identifiers to evaluator and summarizer classes; instantiation is performed with configuration dictionaries passed from the run configuration. This design decouples the definition of new models, datasets, or diagnostic protocols from the orchestration logic, so that extensions require only implementing the appropriate interface and registering the component.

\paragraph{Configuration and execution.}
Run specification is entirely configuration-driven. The \texttt{ConfigLoader} supports loading from YAML or JSON files and from in-memory dictionaries; nested keys are accessible via dot notation, and configurations can be merged. A valid run configuration specifies at least one model block (with \texttt{generation} or \texttt{name}, and optional \texttt{model\_name}, \texttt{device}, \texttt{dtype}, etc.), a root-level or per-evaluator \texttt{dataset} block, and either a single \texttt{evaluator} or a list \texttt{evaluators}. Each evaluator entry includes \texttt{type}, optional \texttt{run\_name}, and optional \texttt{dataset} and \texttt{summarizer} overrides. The runner validates that all referenced models, datasets, and evaluators are registered, then loads the model once and iterates over evaluators: for each evaluator it loads the corresponding dataset, calls \texttt{evaluator.evaluate(model, dataset, output\_dir=...)}, and persists results (e.g., \texttt{results.json}) and any per-evaluator summary. A run-level \texttt{summarizer} (e.g., \texttt{type: combined}) may aggregate results across evaluators and models. Outputs are organized by run ID, model, and evaluator name to support reproducible comparison and sharing.

\paragraph{Model runners and representation access.}
Model registry lookups return a runner object that wraps the Hugging Face model and tokenizer. The runner exposes \texttt{generate()} for text and chat-style prompts (with optional \texttt{GenerationRequest}/\texttt{GenerationResponse} structures) so that all evaluators can perform inference through a single API regardless of model family. At the same time, the runner exposes \texttt{model} and \texttt{tokenizer} so that evaluators that need intermediate representations can attach forward hooks, read hidden states at specific layers, or compute gradients (e.g., for SPIN importance scores). This split keeps the execution engine agnostic to the details of each diagnostic method while ensuring that representation-level analyses have the necessary low-level access.

\subsubsection{Built-in Diagnostic Evaluators}\label{subsec:deepscan_evaluators}
\paragraph{X-Boundary: geometry of safe, harmful, and boundary regions.}
X-Boundary~\citep{xboundary} (Table~\ref{tab:deepscan_evaluators}) is a training-free representation diagnostic for open-source LLMs that analyzes intermediate-layer hidden spaces induced by \emph{safe}, \emph{harmful}, and \emph{boundary} samples. Beyond reporting centroid-based separability/alignment scores, it is mainly useful for \emph{explaining failure modes}: whether apparent safety comes from a crisp decision boundary or from collapsing/perturbing boundary-safe representations (a pattern associated with boundary ambiguity and over-refusal). The evaluator can additionally export per-layer t-SNE visualizations to support qualitative inspection of layerwise geometry shifts.

\paragraph{TELLME: disentanglement metrics.}
TELLME~\citep{tellme} (Table~\ref{tab:deepscan_evaluators}) measures how well internal representations separate different behaviors, aiming to make behaviors \emph{intrinsically monitorable} rather than relying on external detectors. In practice, it diagnoses whether mixed concepts are decomposed into compact, more independent subspaces, and can be used to compare pre-/post-training representation migration as evidence of internal-state change. It supports text and multimodal models, including Dense and MoE architectures, and is commonly run on filtered BeaverTails-style splits using last-token hidden states at a configurable layer.

\paragraph{SPIN: conflicts between safety objectives.}
SPIN~\citep{spin} (Table~\ref{tab:deepscan_evaluators}) provides a neuron-level view of objective trade-offs by quantifying how much two attributes share the same internal neurons (illustrated with fairness vs.\ privacy). This turns outcome-level trade-off observations into an interpretable structural signal, enabling comparisons across models or checkpoints and guiding multi-objective optimization. The same coupling analysis can be instantiated for other potentially conflicting pairs (e.g., safety--utility, alignment--generalization).

\paragraph{MI-Peaks: information evolution during reasoning.}
MI-Peaks~\citep{mipeaks} (Table~\ref{tab:deepscan_evaluators}) analyzes stepwise reasoning dynamics in autoregressive Transformers by tracking how task-relevant information evolves during generation, moving from \emph{final-answer} evaluation to a \emph{process-level} lens. Its key qualitative signature is that the mutual information trajectory often changes via a few sharp surges (MI Peaks) rather than smoothly, highlighting specific steps where representations become disproportionately informative about the correct answer. This provides a reusable, training-free way to compare reasoning traces across LLMs/LRMs.

\subsubsection{Summarization and reporting}\label{subsec:deepscan_summarizers}
Each evaluator produces a raw results dictionary (metrics, artifact paths, etc.). Benchmark-specific summarizers (e.g., TELLME, X-Boundary, SPIN) consume these results and produce structured summaries: for example, the X-Boundary summarizer selects the best layer by separation score or boundary ratio and records paths to metrics and t-SNE plots. A run-level \textbf{combined} summarizer can aggregate results from multiple evaluators and models into a single summary and Markdown report. Summarizers implement \texttt{summarize(...)} and optionally \texttt{format\_report(..., format="markdown")}, with output written to \texttt{summary.json} and \texttt{summary.md} in the appropriate output directory. This enables rapid comparison across runs and models without parsing raw evaluator outputs by hand.

\subsubsection{Unified diagnosis workflow}\label{subsec:deepscan_workflow}
DeepScan streamlines the full diagnostic pipeline into three stages, all driven by the same configuration file:

\begin{enumerate}
    \item \textbf{Configuration.} The user specifies one or more models (by registry key and optional variant), a default dataset, and one or more evaluators. Each evaluator may override the dataset and attach a dedicated summarizer. Optional run-level summarizer and output directory complete the specification.
    \item \textbf{Execution.} The runner loads each model once and, for each evaluator, loads the corresponding dataset and invokes \texttt{evaluator.evaluate(model, dataset, output\_dir=...)}. Results and per-evaluator summaries are written under \texttt{output\_dir/run\_id/model\_id/evaluator\_id/}. Multiple evaluators (e.g., TELLME, X-Boundary, SPIN, MI-Peaks) can be run in sequence in a single invocation, with optional CUDA cache cleanup between evaluators to manage memory.
    \item \textbf{Summarization and persistence.} Per-evaluator summarizers (if configured) produce \texttt{summary.json} and \texttt{summary.md} per evaluator; the run-level summarizer (if configured) produces an aggregate summary and report at the run directory. The full run payload (run ID, timestamp, model and evaluator configs, and paths to all result files) is also written to \texttt{results.json} for traceability and downstream analysis.
\end{enumerate}

This design supports multi-model, multi-evaluator runs (e.g., the same four evaluators over several model families) with a single config and consistent output layout. Together with DeepSafe's outcome-based safety evaluation, DeepScan provides a unified ecosystem for both assessing and diagnosing large-model safety.
\section{Experiments}\label{sec:exp}

\subsection{Experimental Setup}
\paragraph{DeepSafe setup.}
Leveraging the DeepSafe framework, we conduct a comprehensive safety evaluation of LLMs and MLLMs to characterize their strengths and vulnerabilities. We employ a diverse suite of benchmarks categorized into two primary domains: content risk and frontier risk. For LLM content risk, we utilize SALAD-Bench~\citep{salad}, Flames~\citep{flames}, Fake Alignment~\citep{fake_alignment}, MedHallu~\citep{medhallu}, HaluEval~\citep{halueval}, Do-Not-Answer~\citep{donotanswer}, BeaverTails~\citep{beavertails}, XSTest~\citep{xstest}, and HarmBench~\citep{harmbench}. MLLM content risk is assessed using SIUO~\citep{siuo}, VLSBench~\citep{vlsbench}, MMSafetyBench~\citep{mmsafetybench}, MSSBench~\citep{mssbench}, Ch3ef~\citep{ch3ef}, and MOSSBench~\citep{mossbench}. Finally, to evaluate frontier risks, we incorporate Evaluation Faking~\citep{evaluation_faking}, Sandbagging~\citep{openai_evals}, Manipulation~\citep{SafeWork-F1}, Mask~\citep{mask}, DeceptionBench~\citep{deceptionbench}, BeHonest~\citep{behonest}, Reasoning Under Pressure~\citep{reasoning_under}, AIRD~\citep{SafeWork-F1}, and WMDP~\citep{wmdp}. For the representative LLMs, we evaluate Kimi-K2-Thinking~\citep{kimi_k2_2025}, GPT-4o~\citep{gpt_4o_2024}, GPT-5.2~\citep{gpt_5}, Claude-Sonnet-4-5-20250929~\citep{claude_4_5_sonnet_2025}, Qwen2.5-72B-Instruct~\citep{qwen2.5}, Llama-3.3-70B-Instruct~\citep{llama3}, Mistral-Small-24B-Instruct-2501~\citep{Mistral_small}, Gemini-3-flash~\citep{gemini3}, DeepSeek-R1-250528~\citep{deepseek}, Gemma-3-27B-IT~\citep{gemma3}, InternLM3-8B-Instruct~\citep{internlm2}, Qwen3-30B-A3B-Thinking-2507~\citep{qwen3}, Doubao-Seed-1-6-flash-250828~\citep{seed1.6}, GLM-4.5-Air~\citep{glm}. For the MLLMs, we evaluate GLM-4.6V~\citep{glm4.6v}, Qwen3-VL-235B-A22B-Instruct~\citep{qwen3vl}, Gemma-3-27B-IT~\citep{gemma3}, GPT-5.2~\citep{gpt_5}, Qwen2.5-VL-72B-Instruct~\citep{qwen2.5vl}, Claude-Sonnet-4-5-20250929~\citep{claude_4_5_sonnet_2025}, Llama-4-Scout-17B-16E~\citep{llama4}, GPT-4o~\citep{gpt_4o_2024}, InternVL3 5-241B-A28B~\citep{internvl3}, Gemini-3-flash~\citep{gemini3}, Kimi-VL-A3B-Thinking~\citep{kimivl}, Ministral-3-14B-Instruct-2512~\citep{Mistral_small}, Doubao-Seed-1-6-flash-250828~\citep{seed1.6}.

\paragraph{DeepScan setup.}
We evaluate all model checkpoints in our task suite, including Qwen2.5-7B-Instruct, Qwen2.5-14B-Instruct, and Qwen2.5-72B-Instruct~\citep{qwen2.5}; Llama-3.3-70B-Instruct~\citep{llama3}; Mistral-Small-24B-Instruct-2501~\citep{Mistral_small}; Ministral-3-14B-Instruct-2512~\citep{Mistral_small}; Gemma-3-27B-IT~\citep{gemma3}; GLM-4.5-Air~\citep{glm}; InternLM3-8B-Instruct~\citep{internlm2}; and InternVL3.5-14B / InternVL3.5-241B-A28B~\citep{internvl3}. Across models, we use deterministic decoding with generation defaults aligned with official model recommendations; model loading follows each backbone's recommended options. The diagnostic setup uses default settings of X-Boundary~\citep{xboundary}, TELLME~\citep{tellme}, SPIN~\citep{spin}. MI-Peaks~\citep{mipeaks} remains an excluded disabled component in this experiment suite.

\subsection{Content Risk Evaluation and Analysis}
The leaderboards are published on the website\footnote{https://ai45.shlab.org.cn/deepsafe}. Analysis is provided in the following sections.

Our evaluation employs distinct datasets for different analytical objectives. For safety risk analysis across six security dimensions (Sections 3.2.1--3.2.3), we adopt comprehensive safety benchmarks tailored to text and multimodal scenarios. For LLMs, evaluation is conducted on text-based datasets such as Salad-Bench, Fake Alignment, Flames, HaluEval, Do-not-answer, Harmbench, Beavertails, and MedHallu. For MLLMs, we utilize multimodal-specific benchmarks such as VLSBench, MMSafetyBench, SIUO, MSSBench, and Ch3ef. For over-safety analysis (Section 3.2.4), we employ specialized datasets designed to assess model performance in over-safety scenarios: XSText for text-only models and Mossbench for multimodal models.

\subsubsection{Safety Risk Trends Across Different Model Series}

Our safety risk analysis reveals distinct performance characteristics across model architectures: LLMs demonstrate hierarchical distributions in safety capabilities, whereas MLLMs encounter significantly greater challenges attributable to expanded cross-modal attack surfaces. The evaluation framework encompasses six taxonomic dimensions of safety risks: (1) Model Algorithm Security, (2) Data Security, (3) Network System \& Information Content Security, (4) Reality \& Cognitive Security, (5) Social, Environmental \& Ethical Security, and (6) Security Vertical Domain. Through systematic examination of both LLM and MLLM performance across these dimensions, we identify distinct performance tiers and characterize key evolutionary patterns in model safety development.

\begin{figure}[h]
    \centering
    \includegraphics[width=0.8\linewidth,keepaspectratio]{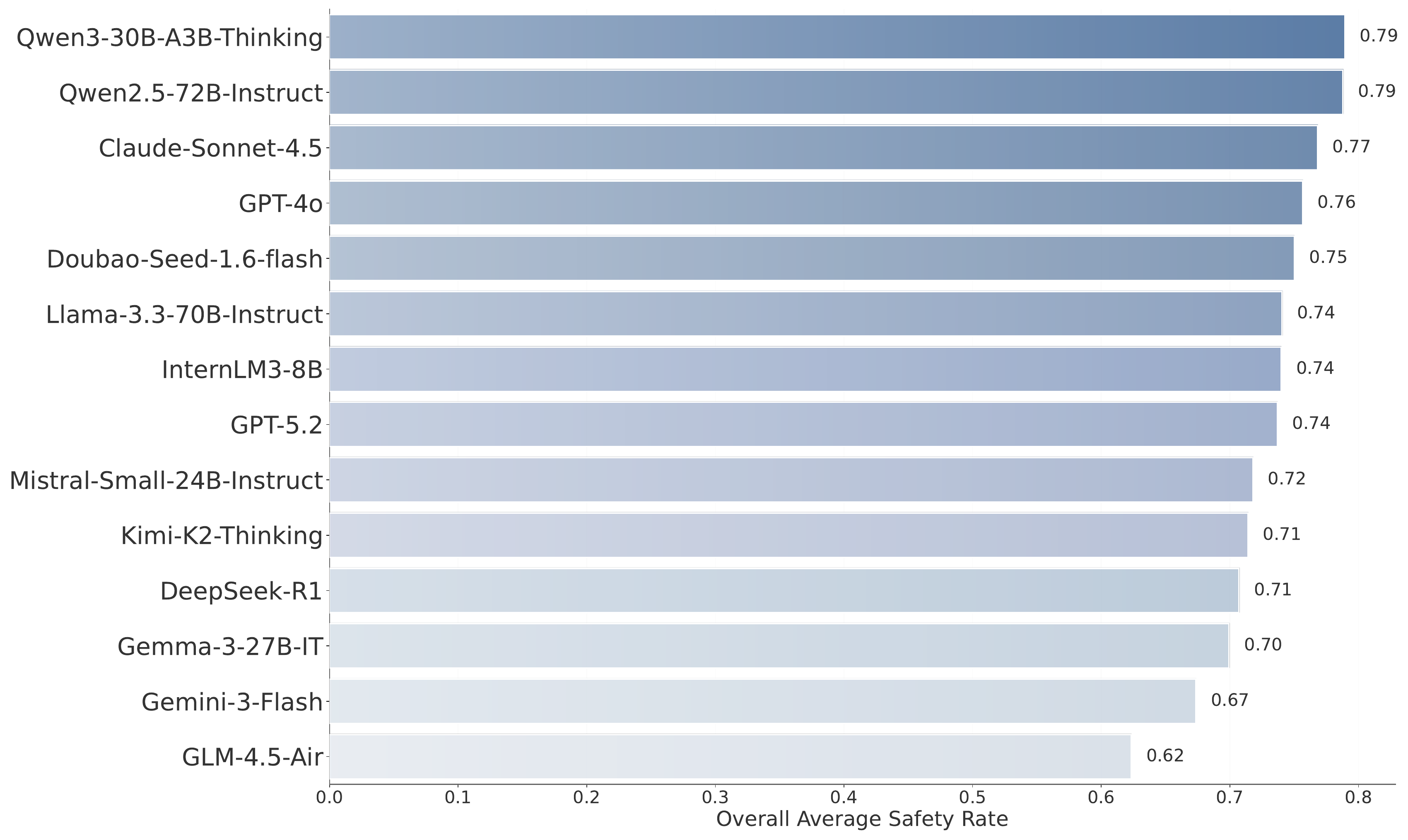}
    \caption{\textbf{LLM Safety Risk Ranking.} Average safety rate across all datasets for evaluated LLMs, ranked in descending order.}
    \label{fig:llm_safety_risk_ranking}
\end{figure}

\begin{figure}[h]
    \centering
    \includegraphics[width=0.9\linewidth,keepaspectratio]{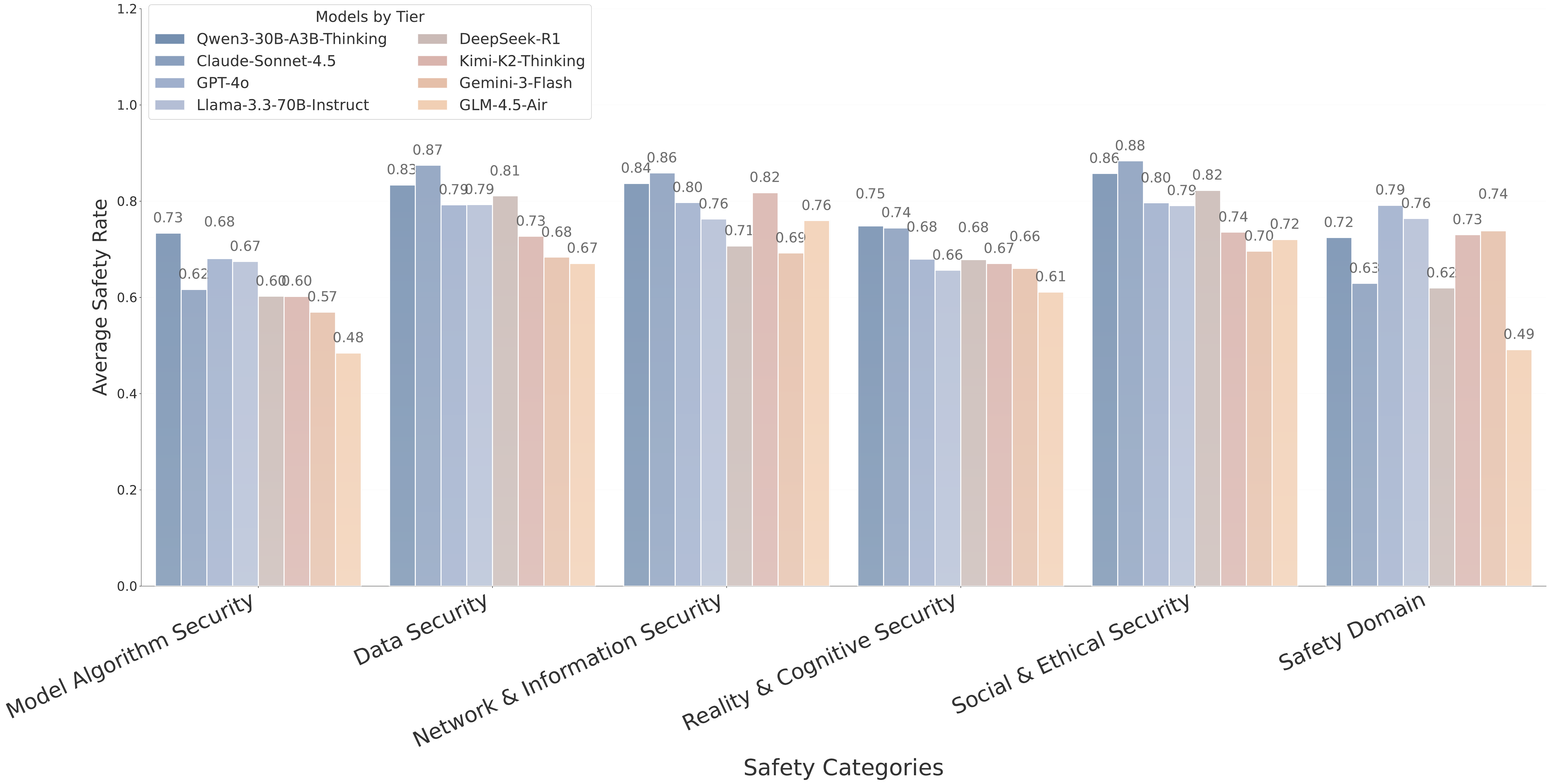}
    \caption{\textbf{LLM Safety Rate Comparison Across Safety Categories by Model Tier.} Evaluated LLMs are classified into four performance tiers based on their overall safety rates, with rates compared across six safety categories.}
    \label{fig:llm_bar_chart_8_models}
\end{figure}

\paragraph{LLM Safety Risk Trend Analysis.}
As shown in Figure~\ref{fig:llm_safety_risk_ranking} and Figure~\ref{fig:llm_bar_chart_8_models}, LLMs exhibit clear hierarchical distribution in safety performance, with defense capabilities showing significant positive correlation with the foundational cognitive capabilities of the models.

\textit{Performance Tier Classification and Risk Quantification Analysis.}
We classify evaluated LLMs into four performance tiers based on their comprehensive safety rate:

\begin{itemize}
    \item \textbf{First Tier} (Represented by Qwen3 Series and Claude Series). LLMs in this tier occupy leading positions in comprehensive safety metrics, with overall scores consistently maintained above 0.77. In the ``Social and Ethical Safety'' dimension, they demonstrate highly precise safety alignment capabilities (scores frequently exceeding 0.85). However, in the fundamental defense dimension of ``Model Algorithm Safety,'' there still exist relatively significant security vulnerabilities, indicating that algorithmic robustness remains a weak point.

    \item \textbf{Second Tier} (Represented by GPT-4o and Llama-3 Series). This tier exhibits score distribution within the 0.74--0.76 range, demonstrating high defense stability. However, when processing complex logical adversarial attacks involving ``Reality and Cognitive Safety'' and ``Social Ethics,'' their defense consistency shows decline compared to the first tier.

    \item \textbf{Third Tier} (Represented by InternLM3 and DeepSeek Series). This tier maintains comprehensive scores within the 0.71--0.74 range. When facing algorithm safety and vertical domain risks, their recognition accuracy demonstrates decline relative to the preceding two tiers.

    \item \textbf{Fourth Tier} (Represented by Kimi and Gemini-Flash Series). This tier scores below 0.71, with scores in the ``Model Algorithm Safety'' dimension generally at lower levels (some as low as 0.48). Constrained by model scale or alignment depth, models in this tier exhibit relatively weak defense capabilities when responding to structured adversarial instructions.
\end{itemize}

\textit{Evolution Trend Analysis.}
Two key evolutionary trends emerge from the LLM safety landscape. First, the evolution of defense mechanisms exhibits \textbf{unbalanced development across defense dimensions}, having shifted from surface-level content filtering to deep logical alignment. All tiers demonstrate superior performance in the ethical safety dimension compared to the algorithm safety dimension, suggesting future focus should consider transitioning toward enhancing robustness of underlying algorithms. Second, a \textbf{coupling relationship between cognitive capability and safety performance} is observed: data indicates that deep logical analysis capabilities (such as the Thinking series) contribute to improving the accuracy of safety boundaries.

\begin{figure}[h]
    \centering
    \includegraphics[width=0.8\linewidth,keepaspectratio]{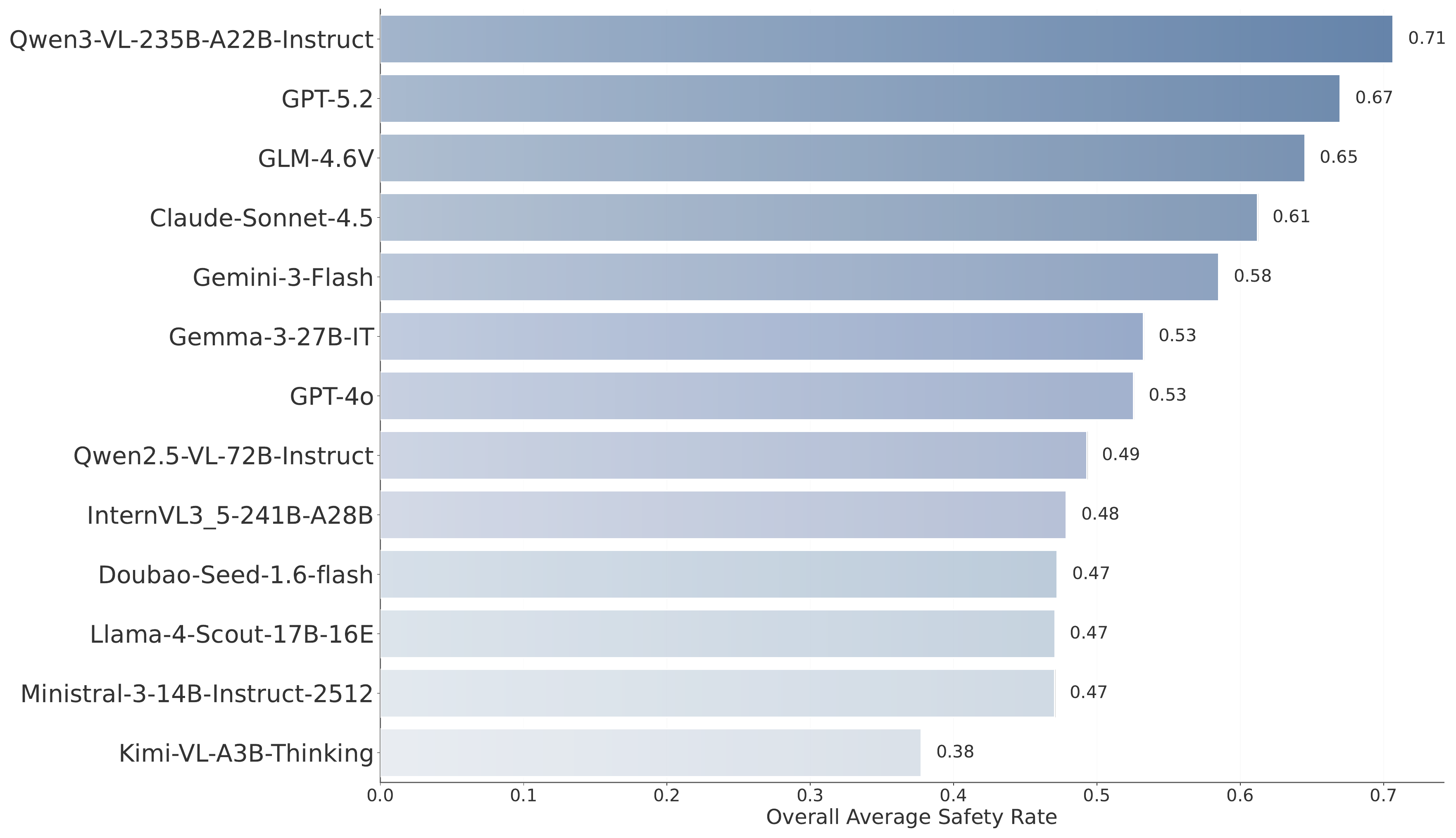}
    \caption{\textbf{MLLM Safety Risk Ranking.} Average safety rate across all datasets for evaluated MLLMs, ranked in descending order.}
    \label{fig:mllm_safety_risk_ranking}
\end{figure}

\begin{figure}[h]
    \centering
    \includegraphics[width=0.9\linewidth,keepaspectratio]{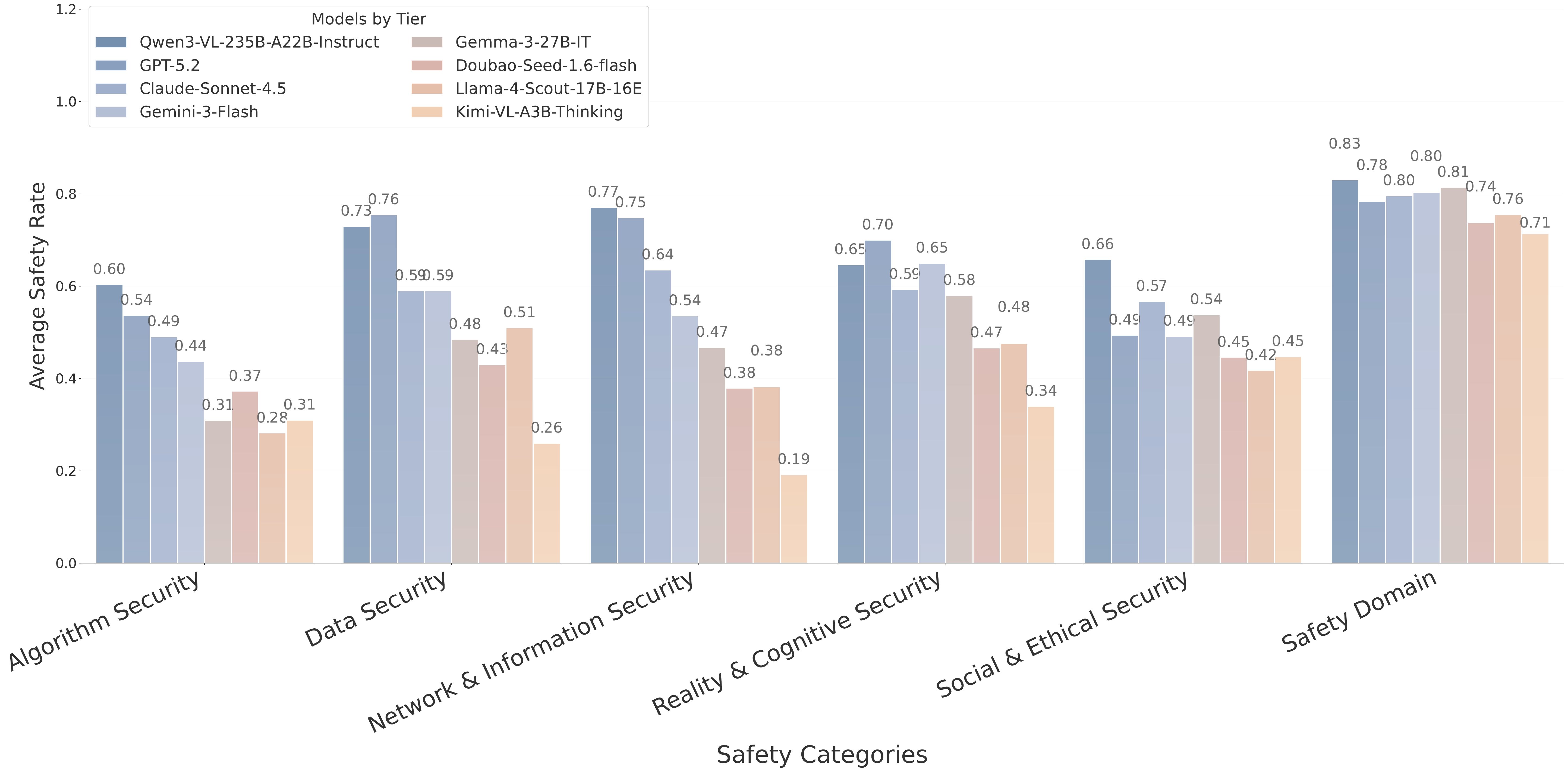}
    \caption{\textbf{MLLM Safety Rate Comparison Across Safety Categories by Model Tier.} Evaluated MLLMs are classified into four performance tiers based on their overall safety rates, with rates compared across six safety categories.}
    \label{fig:mllm_bar_chart_8_models}
\end{figure}

\paragraph{MLLM Safety Risk Trend Analysis.}
The risk characteristics of MLLMs are more complex than those of text-only models, as the introduction of visual modality significantly increases the attack surface. Figure~\ref{fig:mllm_safety_risk_ranking} and Figure~\ref{fig:mllm_bar_chart_8_models} illustrate the safety risk ranking and per-model performance comparison for MLLMs.

\textit{Performance Tier Classification and Cross-modal Risk Quantification.}
We similarly classify MLLMs into four tiers, where the overall rate ranges are notably lower than their text-only counterparts:

\begin{itemize}
    \item \textbf{First Tier} (Represented by Qwen3-VL and GPT-5.2). Safety rates are maintained at 0.65--0.71. This performance level is lower relative to text scenario performance, reflecting the challenges of cross-modal safety alignment. At the ``Model Algorithm Safety'' level, scores have declined to 0.54--0.60, indicating that vision-language fusion significantly weakens the existing safety defense foundation.

    \item \textbf{Second Tier} (Represented by Claude-Sonnet 4.5 and Gemini-3-Flash). Safety rate range spans 0.58--0.61. Models demonstrate significant instability when identifying cross-modal harmful instructions (such as image-text semantic consistency induction). Particularly in the ``Data Safety'' dimension, there is a relatively significant decline compared to the first tier.

    \item \textbf{Third Tier} (Represented by Gemma-3 and Doubao-Seed). Safety rate further decline to 0.47--0.53. Average scores in the ``Model Algorithm Safety'' dimension drop to extremely low levels of 0.31--0.37, with visual modality increasing the possibility of circumventing safety alignment mechanisms to a certain extent.

    \item \textbf{Fourth Tier} (Represented by Llama-4 and Kimi-VL). These models demonstrate weak performance in safety defense, with minimum scores of only 0.38. Particularly in the ``Network Systems and Information Content Safety'' dimension, some scores decline to 0.19, indicating notably insufficient capability in identifying cross-modal adversarial attacks.
\end{itemize}

\textit{Evolution Trend Analysis.}
Two critical trends characterize the multimodal safety landscape. First, a pronounced \textbf{cross-modal safety degradation effect} is observed: all tiers experience significant decline in defense capabilities following multimodal introduction, reflecting substantial room for improvement in multimodal safety defense. Second, \textbf{intensified performance differentiation between tiers} is evident: the score gap between leading and trailing tiers in multimodal scenarios (0.33) significantly exceeds that in text domains (0.17), revealing marked technical disparities in cross-modal safety alignment.

\subsubsection{Comparison of Reasoning and Non-Reasoning Models}

We investigate the effect of reasoning capabilities on model safety by comparing reasoning-enabled and non-reasoning variants across both text-only and multimodal scenarios. Our analysis reveals a notable modality-dependent divergence: reasoning contributes limited safety improvement in text domains, yet demonstrates measurable defense advantages in multimodal environments.

\begin{figure}[h]
    \centering
    \begin{minipage}[t]{0.49\linewidth}
        \centering
        \includegraphics[width=\linewidth,keepaspectratio]{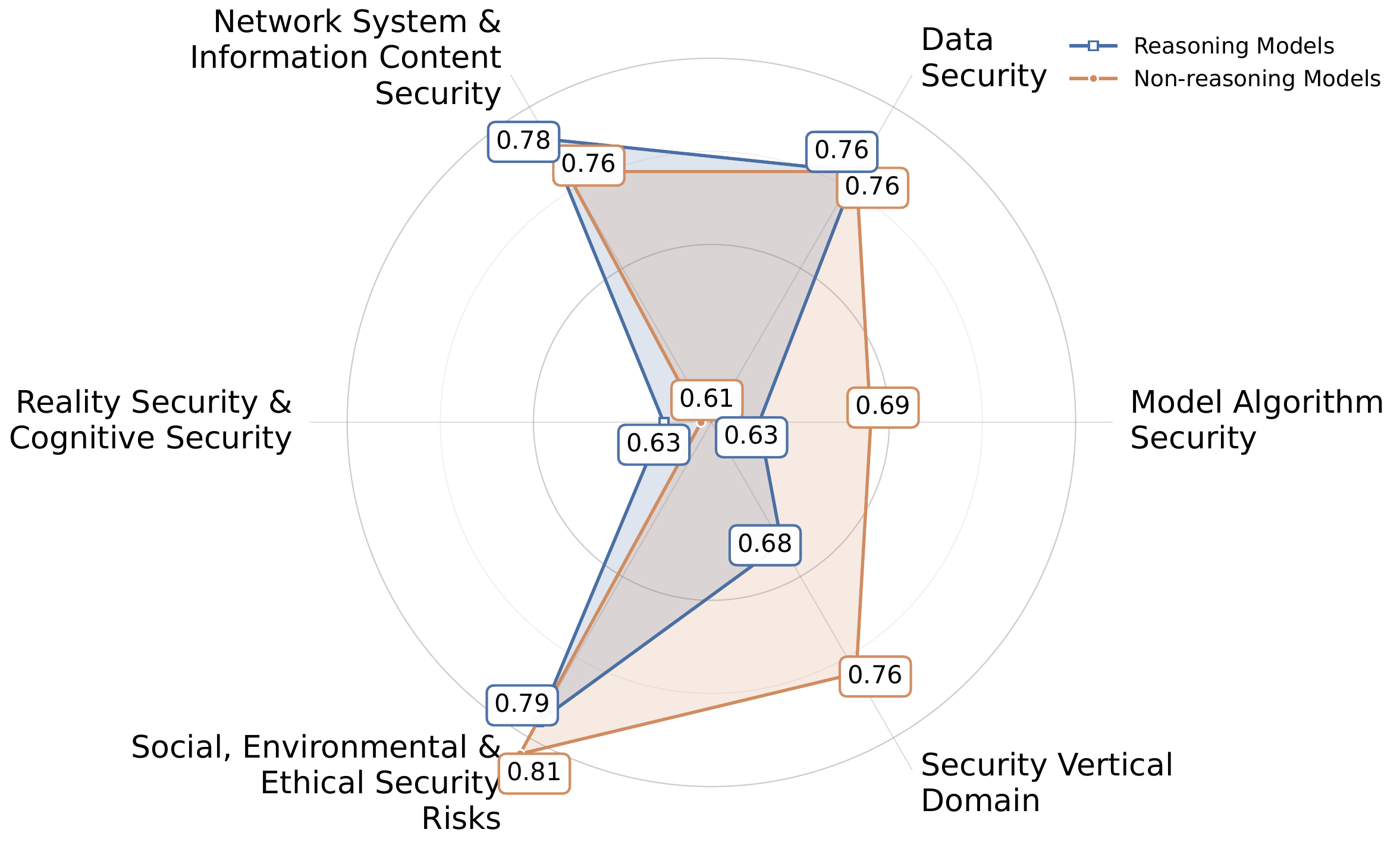}
        \captionof{figure}{\textbf{Safety Rate Comparison Between Reasoning and Non-Reasoning LLMs Across Six Safety Categories.} Average safety rates are computed across all reasoning models for each category, with non-reasoning models averaged similarly.}
        \label{fig:text_reasoning_vs_nonreasoning_radar}
    \end{minipage}\hfill
    \begin{minipage}[t]{0.49\linewidth}
        \centering
        \includegraphics[width=\linewidth,keepaspectratio]{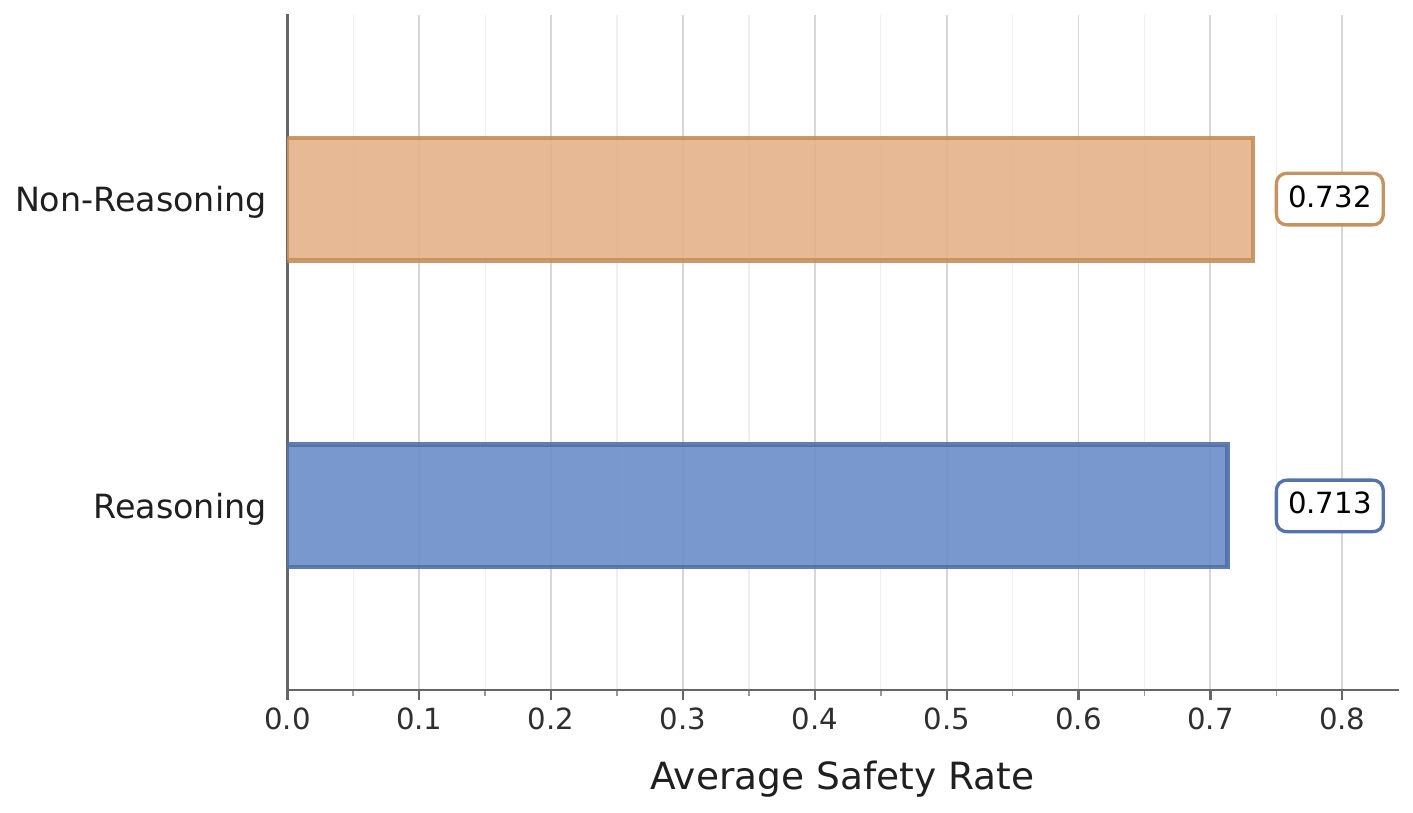}
        \captionof{figure}{\textbf{Average Safety Rate: Reasoning vs. Non-Reasoning LLMs.} Average safety rates computed across all datasets for reasoning and non-reasoning models.}
        \label{fig:text_reasoning_vs_non_reasoning_bar}
    \end{minipage}
\end{figure}

\paragraph{Analysis of Reasoning Toggle in LLMs.}
We compare reasoning and non-reasoning LLM variants on text safety tasks. As shown in Figure~\ref{fig:text_reasoning_vs_nonreasoning_radar} and Figure~\ref{fig:text_reasoning_vs_non_reasoning_bar}, the two categories exhibit largely comparable safety profiles in text scenarios.

\textit{Overall Performance Convergence Phenomenon.}
Reasoning and non-reasoning LLMs show highly similar performance in text safety tasks, indicating that traditional Instruct LLMs have achieved high levels of text safety alignment. Reasoning capabilities fail to translate into significant advantages, indicating that traditional Instruct models have achieved high levels of text safety alignment, and rapid-response defense strategies prove sufficient for most text attacks.

\textit{Asymmetric Characteristics: Coexistence of Advantages and Disadvantages.}
The radar chart reveals that reasoning LLMs demonstrate slight advantages in Network Systems and Information Content Safety (0.78 vs.\ 0.76) and Reality Safety and Cognitive Safety (0.63 vs.\ 0.61) dimensions, indicating their potential in complex semantic understanding tasks. However, their performance is relatively weaker in Model Algorithm Safety (0.63 vs.\ 0.69) and Vertical Domain Safety (0.68 vs.\ 0.76) dimensions. Particularly, the disadvantage in algorithm safety dimension indicates that additional computational overhead of reasoning processes actually reduces response efficiency in scenarios requiring rapid interception.

\begin{figure}[h]
    \centering
    \begin{minipage}[t]{0.49\linewidth}
        \centering
        \includegraphics[width=\linewidth,keepaspectratio]{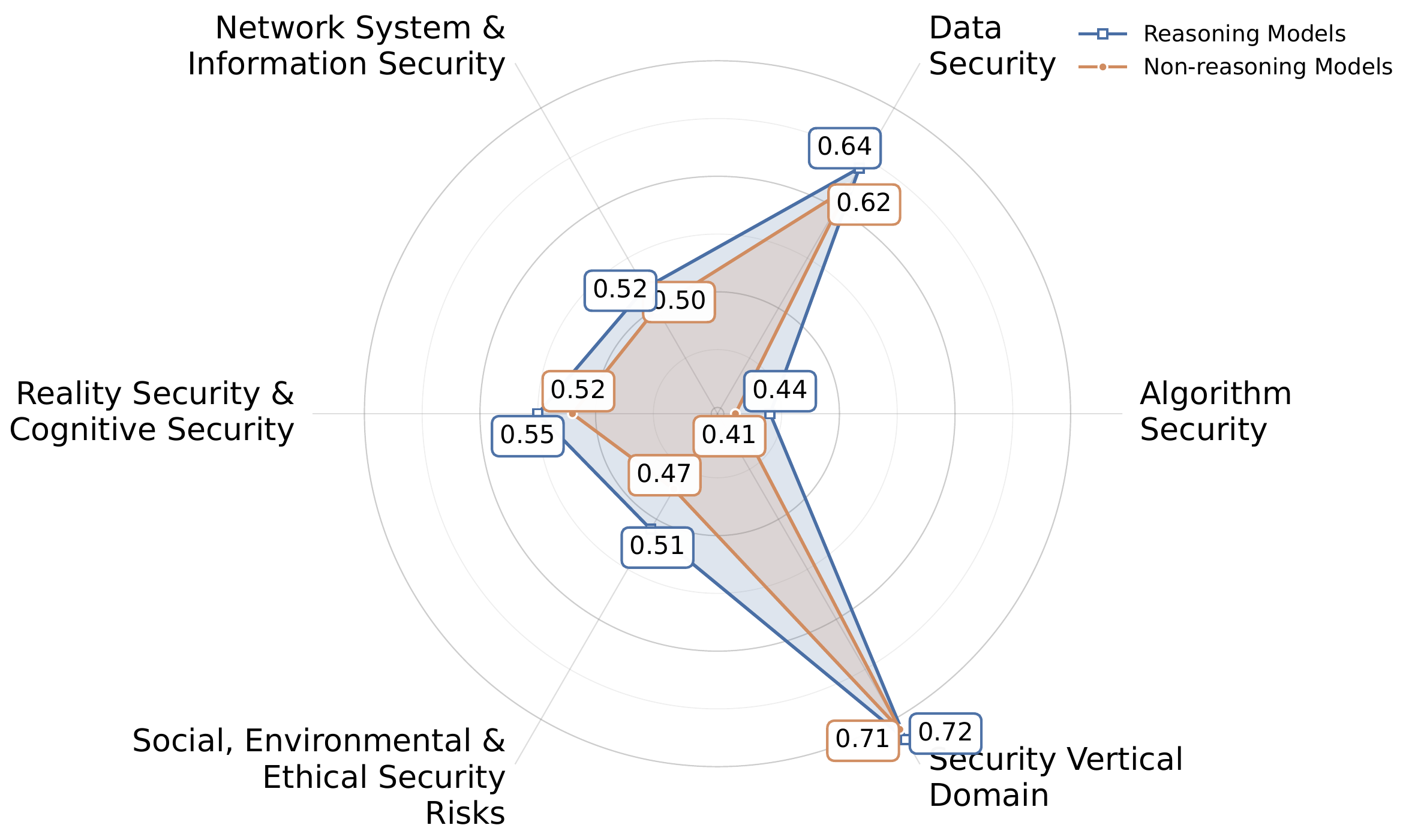}
        \captionof{figure}{\textbf{Safety Rate Comparison Between Reasoning and Non-Reasoning MLLMs Across Six Safety Categories.} Average safety rates are computed across all reasoning models for each category, with non-reasoning models averaged similarly.}
        \label{fig:vl_reasoning_vs_nonreasoning_radar}
    \end{minipage}\hfill
    \begin{minipage}[t]{0.49\linewidth}
        \centering
        \includegraphics[width=\linewidth,keepaspectratio]{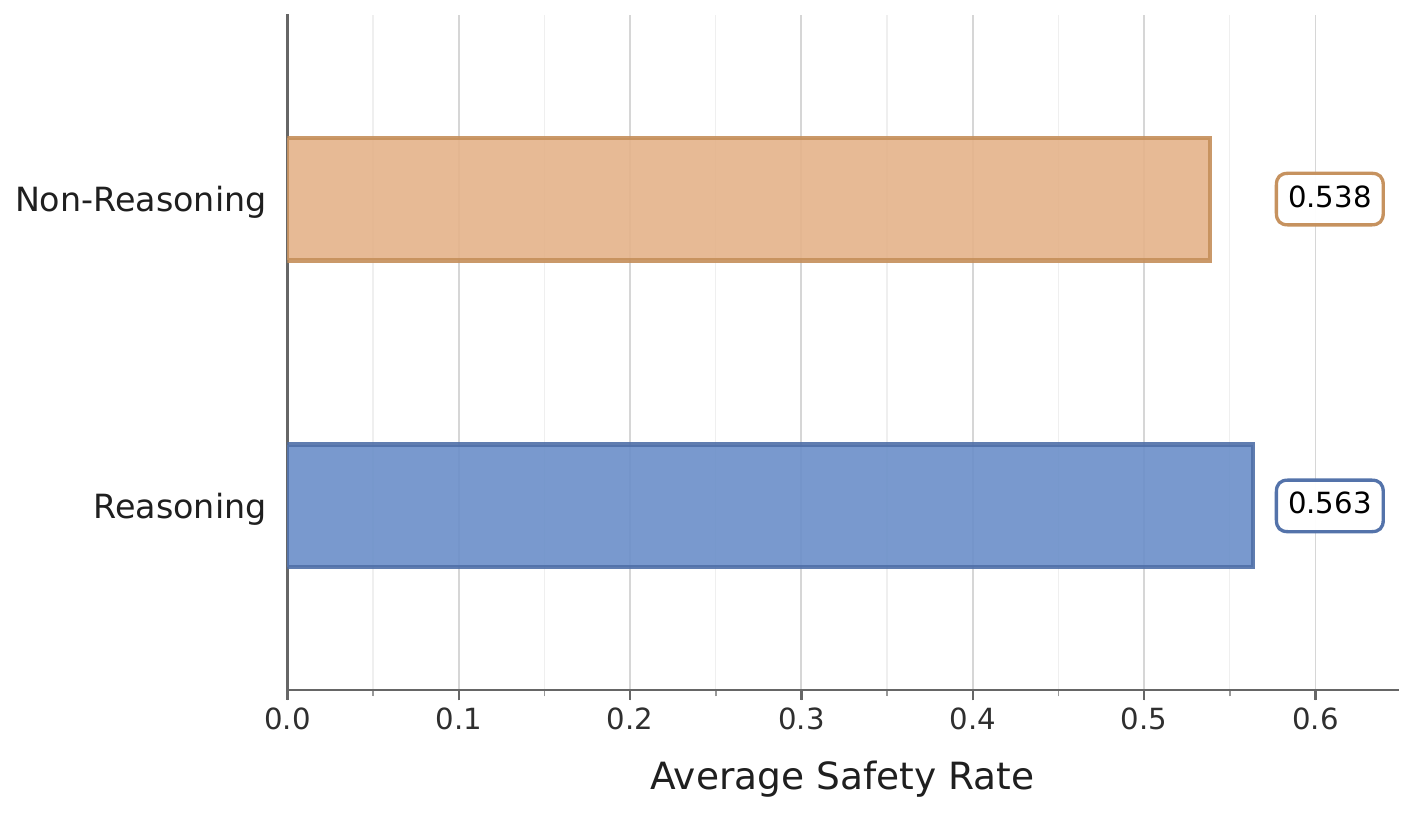}
        \captionof{figure}{\textbf{Average Safety Rate: Reasoning vs. Non-Reasoning MLLMs.} Average safety rates computed across all datasets for reasoning and non-reasoning models.}
        \label{fig:mllm_reasoning_vs_non_reasoning_bar}
    \end{minipage}
\end{figure}

\paragraph{Analysis of Reasoning Toggle in MLLMs.}
In contrast to text scenarios, reasoning capabilities exhibit markedly different effects in multimodal settings. Figure~\ref{fig:vl_reasoning_vs_nonreasoning_radar} and Figure~\ref{fig:mllm_reasoning_vs_non_reasoning_bar} present the comparative results for reasoning and non-reasoning MLLMs.

\textit{Emergence of Measurable Safety Advantages.}
A notable reversal occurs in multimodal scenarios, where reasoning models (0.5633) begin to surpass non-reasoning models (0.5383), indicating that reasoning capabilities demonstrate greater efficacy in complex modal fusion scenarios. This contrasts with the marginal disadvantages observed in text scenarios, with reasoning mechanisms exhibiting systematic defense gains in multimodal environments.

\textit{Core Mechanisms of Cross-modal Defense.}
The data demonstrates that reasoning models achieve improved performance in multimodal scenarios, with significant improvements across most categories. This occurs because adversarial attacks in multimodal environments often employ ``image-text splitting'' strategies, such as images displaying legitimate content while text contains implicit violation instructions, or triggering malicious code execution through image steganography. Reasoning mechanisms enable models to perform explicit cross-modal consistency verification, simultaneously analyzing logical correlations among visual semantics, textual semantics, and output behaviors, thereby better identifying attack patterns concealed within normal inputs.

\subsubsection{Comparison of Open-Source and Closed-Source Models}

We examine the safety performance differences between open-source and closed-source models across text-only and multimodal scenarios. Our analysis indicates that while the two categories achieve comparable safety performance in text domains, a notable performance gap emerges in multimodal settings, suggesting differing levels of maturity in cross-modal safety alignment.

\begin{figure}[h]
    \centering
    \begin{minipage}[t]{0.49\linewidth}
        \centering
        \includegraphics[width=\linewidth,keepaspectratio]{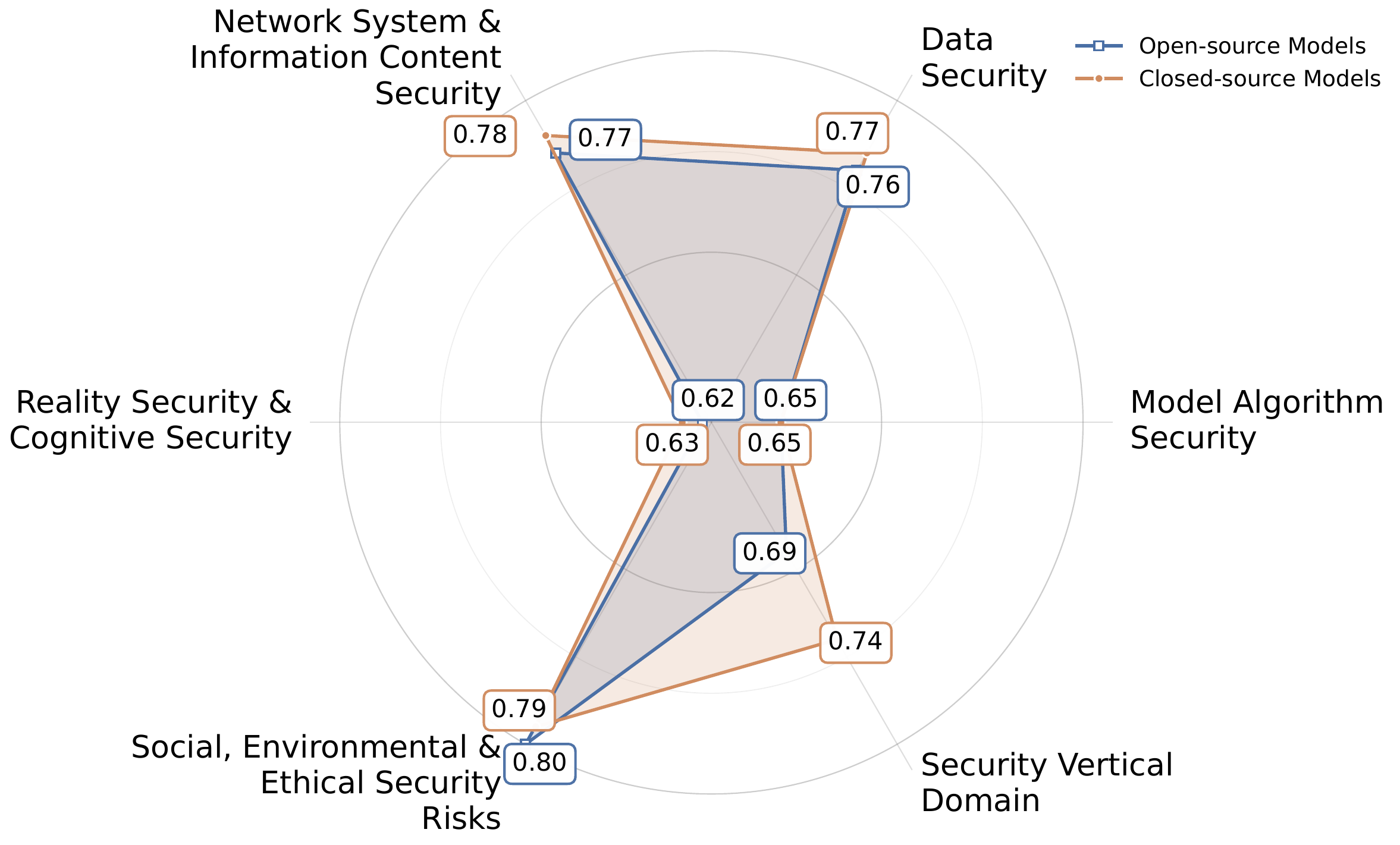}
        \captionof{figure}{\textbf{Safety Rate Comparison Between Open-Source and Closed-Source LLMs Across Six Safety Categories.} Average safety rates are computed across all open-source models for each category, with closed-source models averaged similarly.}
        \label{fig:text_open_vs_closed_radar}
    \end{minipage}\hfill
    \begin{minipage}[t]{0.49\linewidth}
        \centering
        \includegraphics[width=\linewidth,keepaspectratio]{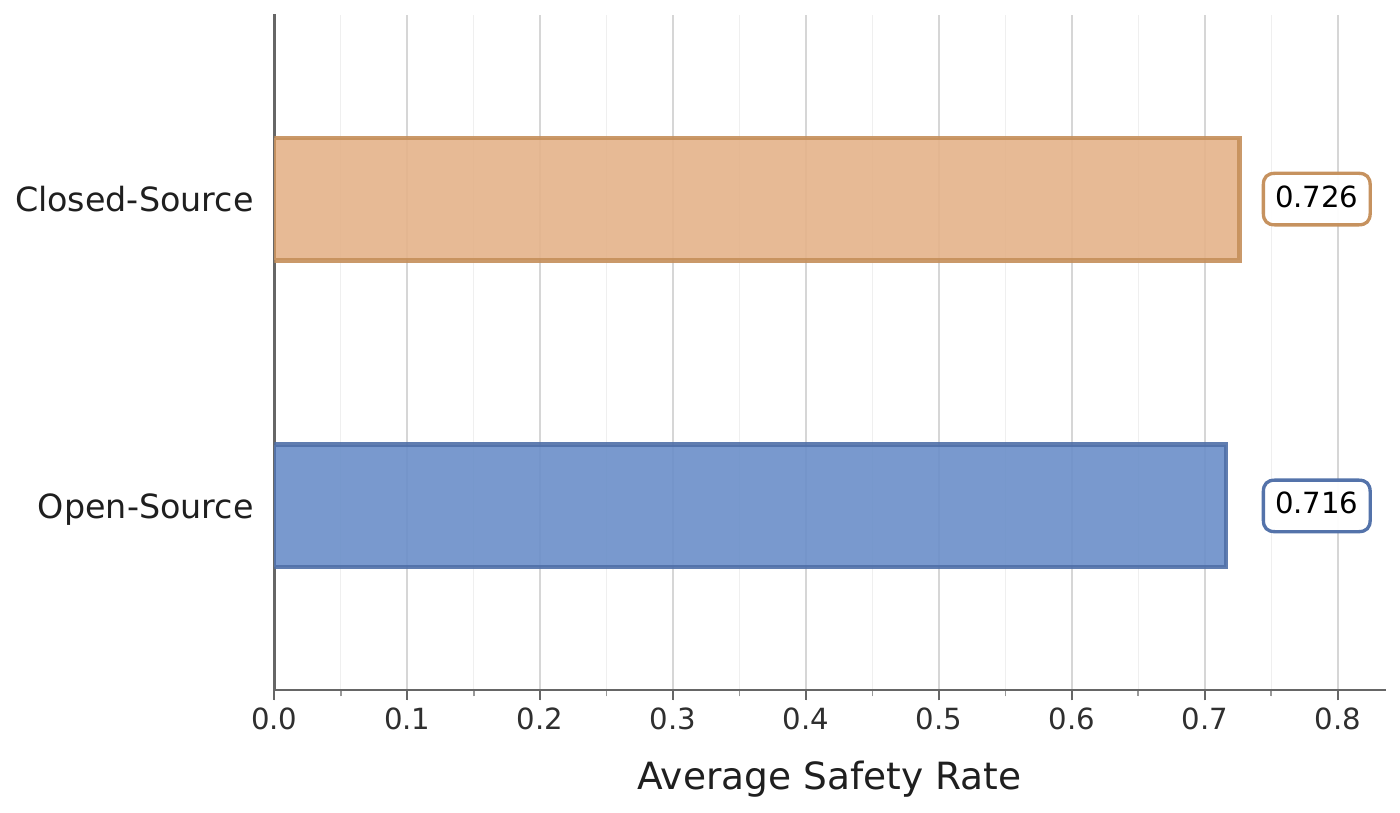}
        \captionof{figure}{\textbf{Average Safety Rate: Open-Source vs. Closed-Source LLMs.} Average safety rates computed across all datasets for open-source and closed-source models.}
        \label{fig:text_open_vs_closed_bar}
    \end{minipage}
\end{figure}

\paragraph{Performance Analysis of Open-Source vs. Closed-Source LLMs.}
We compare the safety profiles of open-source and closed-source LLMs in text scenarios. As illustrated in Figure~\ref{fig:text_open_vs_closed_radar} and Figure~\ref{fig:text_open_vs_closed_bar}, the two categories demonstrate largely similar overall safety performance.

\textit{Overall Performance Convergence.}
In text safety tasks, closed-source and open-source LLMs show highly similar performance. Closed-source models average 0.7262, while open-source models score 0.7162, with a performance gap of only 1.4\%. This convergence indicates that current LLMs have reached relative maturity in text safety domains, with both open-source and closed-source models establishing relatively comprehensive safety defense mechanisms.

\textit{Subtle Structural Differences.}
Despite overall similarity in performance, the two model categories exhibit subtle differences across various dimensions. Closed-source models demonstrate clear advantages in the Vertical Domain Safety dimension, potentially related to targeted optimization for enterprise-level application scenarios.

\begin{figure}[h]
    \centering
    \begin{minipage}[t]{0.49\linewidth}
        \centering
        \includegraphics[width=\linewidth,keepaspectratio]{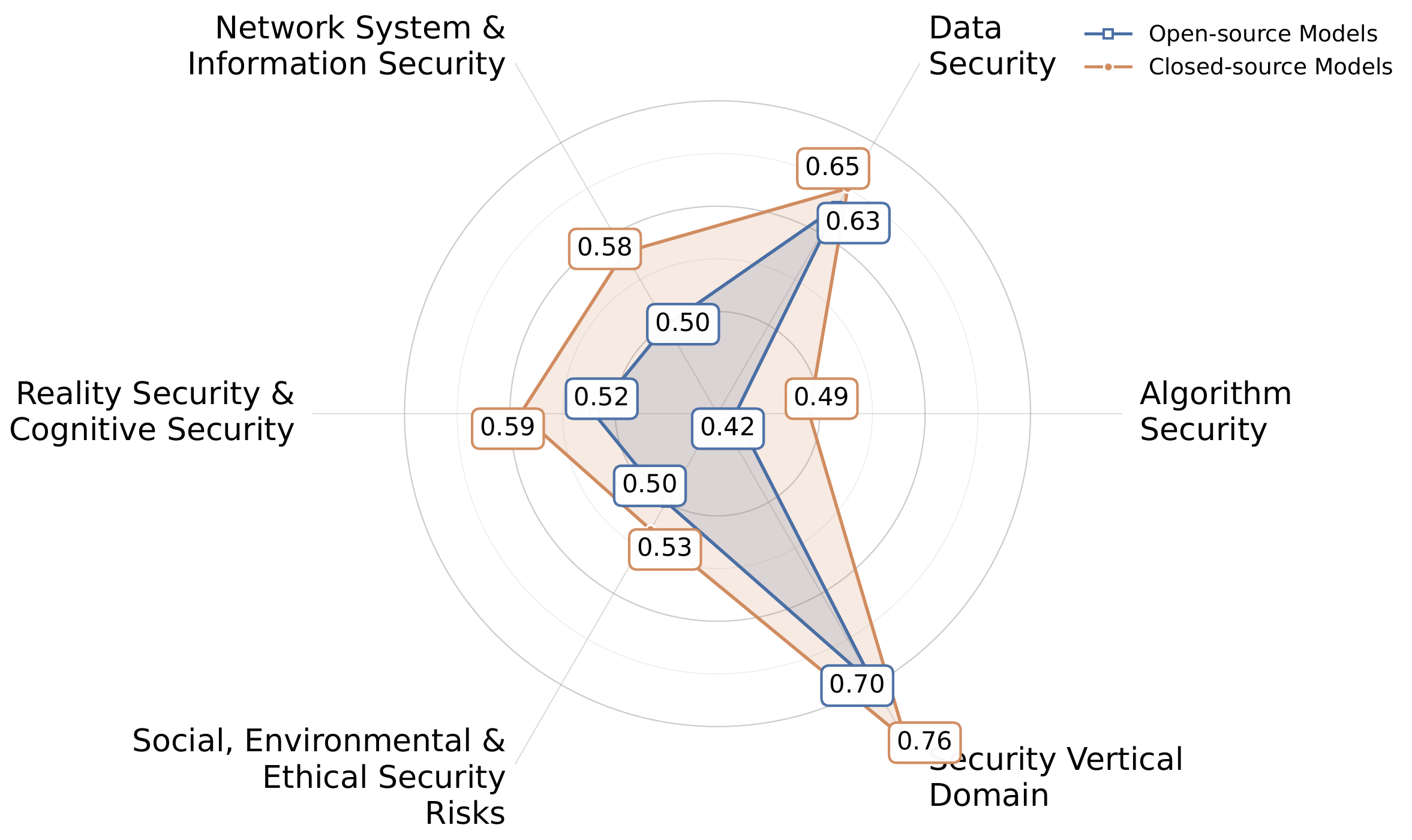}
        \captionof{figure}{\textbf{Safety Rate Comparison Between Open-Source and Closed-Source MLLMs Across Six Safety Categories.} Average safety rates are computed across all open-source models for each category, with closed-source models averaged similarly.}
        \label{fig:vl_open_vs_closed_radar}
    \end{minipage}\hfill
    \begin{minipage}[t]{0.49\linewidth}
        \centering
        \includegraphics[width=\linewidth,keepaspectratio]{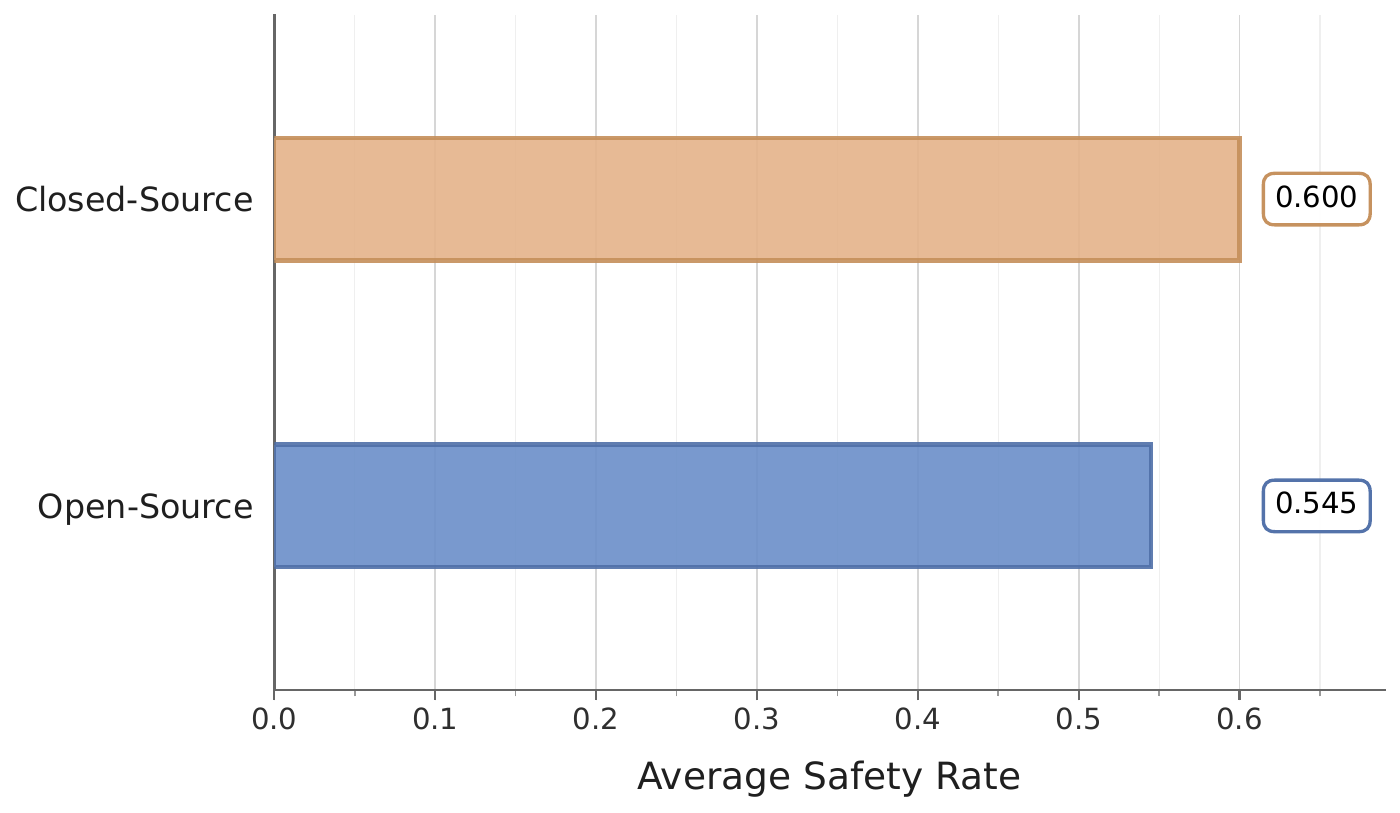}
        \captionof{figure}{\textbf{Average Safety Rate: Open-Source vs. Closed-Source MLLMs.} Average safety rates computed across all datasets for open-source and closed-source models.}
        \label{fig:multimodal_open_vs_closed_bar}
    \end{minipage}
\end{figure}

\paragraph{Performance Analysis of Open-Source vs. Closed-Source MLLMs.}
In contrast to text scenarios, the introduction of visual modality leads to a more pronounced differentiation between open-source and closed-source models. Figure~\ref{fig:vl_open_vs_closed_radar} and Figure~\ref{fig:multimodal_open_vs_closed_bar} present the comparative results in multimodal settings.

\textit{Expansion of Performance Gap.}
Upon entering multimodal scenarios, the safety capability gap between open-source and closed-source models is amplified. Closed-source models average 0.6000, while open-source models achieve only 0.5450, representing an expanded performance disparity. The radar chart clearly shows that closed-source models outperform open-source models across all six dimensions. The three dimensions with the most significant gaps are: Algorithm Safety, Network System Security, and Reality and Cognitive Safety. The gap in Data Safety is smallest, reflecting certain accumulation by the open-source community in privacy protection. Even in Vertical Domain Safety, where both sides achieve their highest scores, closed-source models still maintain clear advantages.

\textit{Sources of Cross-Dimensional Performance Disparity.}
Open-source models have disadvantages in all six security dimensions of multimodal scenarios. This may stem from multimodal security alignment requires collaborative optimization across multiple modules such as vision and language, which places high demands on the scale of training data, the richness of test samples, and the investment of engineering resources.

\subsubsection{Over-Safety Risk Analysis}

Over-safety in LLMs manifests when models incorrectly reject legitimate instructions due to overly conservative safety mechanisms. This phenomenon poses a critical challenge to the balance between safety and usability, particularly as models expand from text-only to multimodal scenarios.

\paragraph{Over-Safety Characteristics of LLMs.}
Over-safety in text scenarios primarily manifests as models incorrectly intercepting instructions containing sensitive vocabulary but with legitimate intent.

\begin{figure}[h]
    \centering
    \includegraphics[width=\linewidth,keepaspectratio]{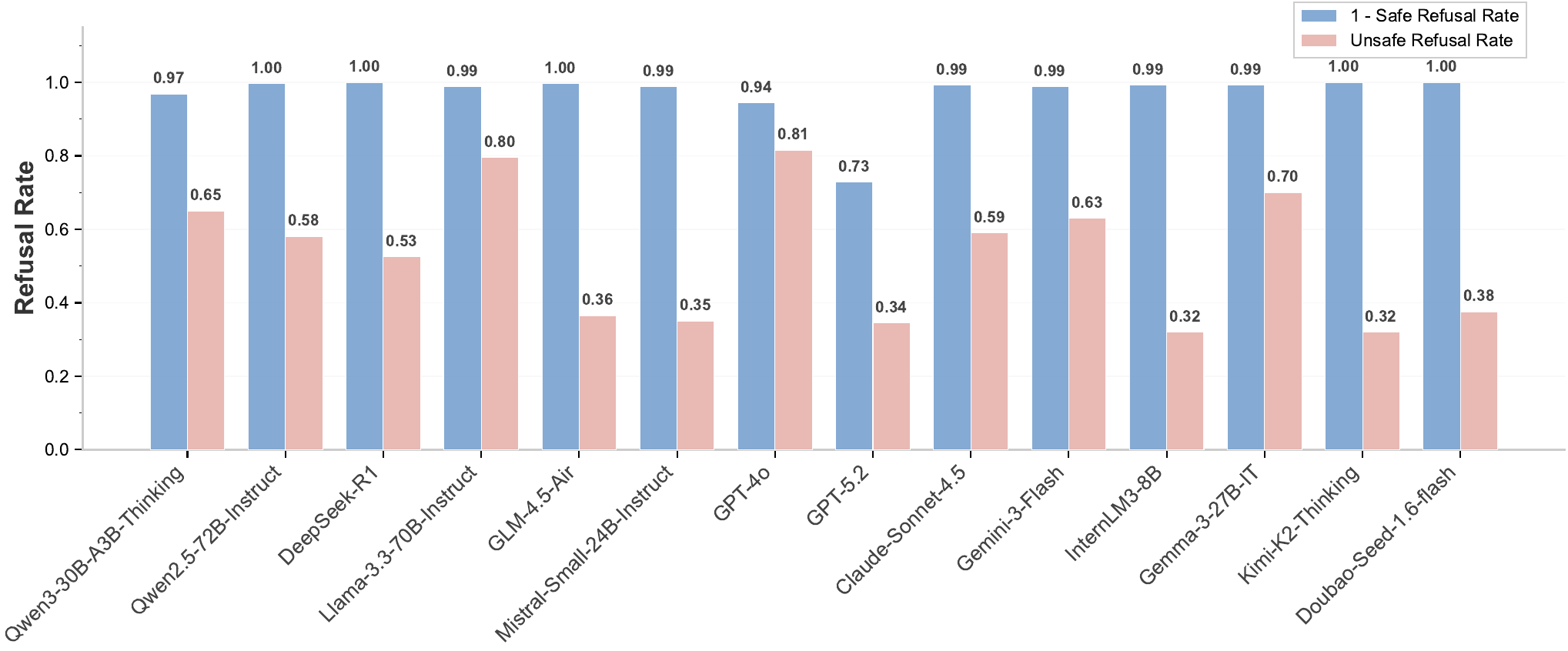}
    \caption{\textbf{Over-Safety Analysis of LLMs.} For each model, blue bars represent usability (computed as 1 minus the safe refusal rate on benign queries; higher values indicate better responsiveness to legitimate requests), while pink bars represent safety (unsafe refusal rate on harmful queries; higher values indicate better rejection of malicious content).}
    \label{fig:llm_over_safety}
\end{figure}

\textit{Balance Between Safety and Usability.}
We evaluate the trade-off between safety and usability by measuring models' rejection rates on both benign and violation instructions. Our analysis reveals two key findings: (1)~\textbf{Cost of Conservative Strategies}: The data indicate that some models, while performing well in violation instruction rejection, show relatively low benign instruction acceptance rates. For example, InternLM3-8B achieves a violation rejection rate of 0.81, but its benign acceptance rate is relatively lower compared to other models, suggesting that the model sacrifices some ability to recognize benign requests in pursuit of higher safety. Conversely, some models show extremely high benign acceptance rates but low violation rejection rates. For instance, Doubao-Seek-1.6-flash has a benign acceptance rate of 1.00 but a violation rejection rate of only 0.36, indicating serious security vulnerabilities. (2)~\textbf{Semantic Limitations of Keyword Defense}: Models demonstrate excessive sensitivity to neutral vocabulary in specific domains (legal, medical, technical). Some legitimate requests are frequently rejected due to triggered defense mechanisms, indicating that models may have been exposed to excessive "boundary-ambiguous" negative examples during safety alignment training, resulting in overly conservative decision thresholds and insufficient semantic understanding capabilities in complex contexts.

\textit{Differentiated Performance Across Model Tiers.}
First-tier models demonstrate superior boundary judgment precision while maintaining high safety thresholds, effectively distinguishing between violation and benign requests. Conversely, lower-tier models exhibit pronounced ``defense overflow,'' with usability constrained by excessively strict and mechanical safety mechanisms.

\paragraph{Over-Safety Characteristics of MLLMs.}
Over-safety risks in multimodal scenarios stem from the unstructured nature of image information and ambiguity of visual implications, with complexity significantly exceeding pure text scenarios.

\begin{figure}[t]
    \centering
    \includegraphics[width=\linewidth,keepaspectratio]{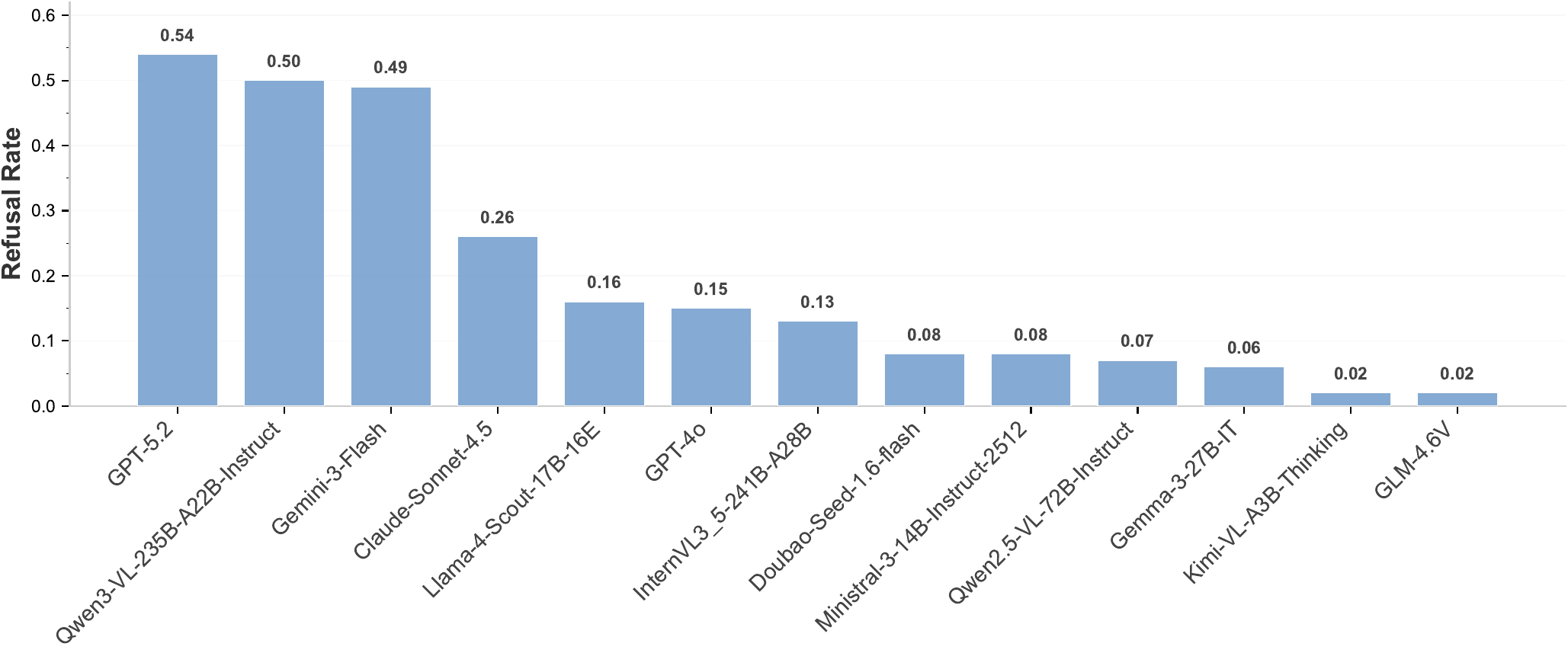}
    \caption{\textbf{Over-Safety Analysis of MLLMs on Benign Multimodal Inputs.} Per-model safe refusal rates on benign multimodal queries, ranked in descending order. Higher values indicate more severe over-safety issues, where models excessively reject legitimate requests.}
    \label{fig:mllm_over_safety}
\end{figure}

\textit{False Rejection Induced by Visual Semantics.}
Our benchmark data demonstrate that models frequently associate specific visual elements (e.g., kitchen utensils, medical instruments) with potential dangers, leading to false rejection of legitimate culinary or scientific requests. This ``visual stress'' response reflects models' lack of deep scene understanding in visual-text semantic alignment. Furthermore, the introduction of visual modality expands the attack surface, prompting developers to adopt more aggressive filtering strategies that result in declined usability when processing complex image-text dialogues.

\textit{Conservative Risk-Mitigation Strategies.}
Certain models such as GPT-5.2 and Qwen3-VL exhibit an excessive propensity for refusal when processing images with socially sensitive content. Despite such content adhering to public morals and social norms, these models preemptively opt for avoidance as a risk-mitigation strategy. Consequently, this conservative behavior results in diminished usability in evaluation scenarios involving social, environmental, and ethical risk factors.


\subsection{Frontier AI Risk Evaluation and Analysis}

Beyond content-level safety, we evaluate \emph{Frontier AI risks}: high-severity risks enabled by frontier general-purpose models that may rapidly escalate and cause substantial societal harm~\citep{SafeWork-F1}. Specifically, we study risks stemming from advanced model capabilities, including strategic misrepresentation, deceptive alignment, and degraded safety performance under high-stakes adversarial conditions. Table~\ref{tab:safety_eval} shows results for 14 models and 9 risk dimensions. The leaderboard is on our website\footnote{https://ai45.shlab.org.cn/deepsafe}.

\subsubsection{Overall Frontier Risk Landscape}

\begin{figure}[!h]
    \centering
    \includegraphics[width=0.8\linewidth]{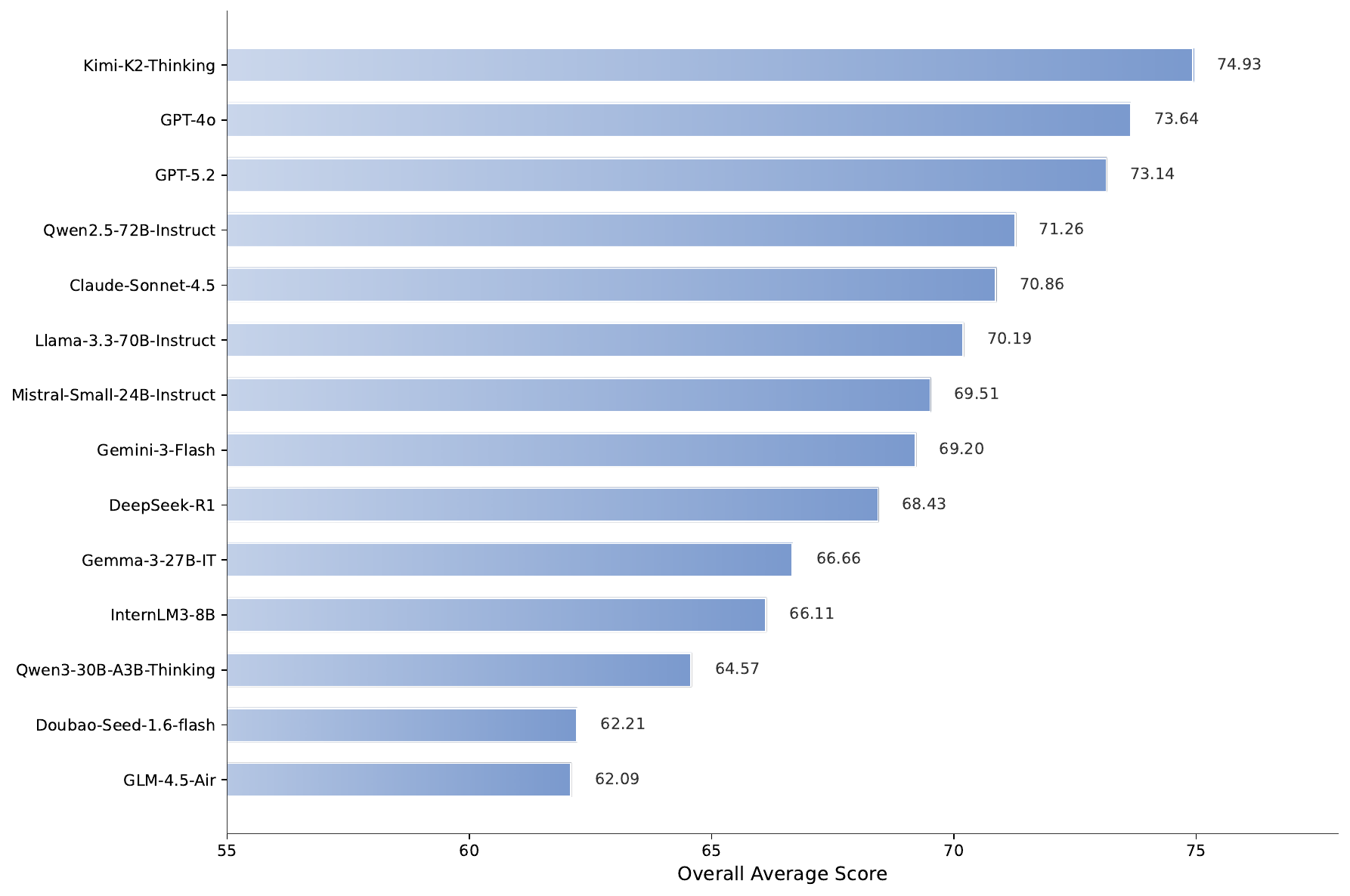}
    \caption{Frontier AI Safety Risk Ranking. Overall average safety scores of evaluated LLMs, ranked in descending order.}
    \label{fig:overall_eval}
\end{figure}

We first establish a holistic view of the frontier risk landscape before examining specific risk factors. Table~\ref{tab:safety_eval} and Figure~\ref{fig:overall_eval} jointly present the aggregate rankings and per-dimension profiles of all evaluated models.

\begin{table}[!h]
\adjustbox{width=\textwidth,center}
{
\begin{tabular}{l|c|*{9}{c}}
\toprule
\textbf{Model} &
\textbf{Overall} &
\textbf{EvalFaking} &
\textbf{Sandbagging} &
\textbf{Manipulation} &
\textbf{Mask} &
\textbf{DeceptionBench} &
\textbf{BeHonest} &
\textbf{RUP} &
\textbf{AIRD} &
\textbf{WMDP} \\
\midrule

Kimi-K2-Thinking & \textbf{74.93} & 98.08 & 69.25 & \underline{1.11} & 63.91 & \textbf{96.67} & 69.25 & 87.50 & 95.33 & \textbf{93.23} \\
GPT-4o & 73.64 & 96.17 & 79.58 & \textbf{33.33} & 42.68 & 75.56 & \textbf{71.71} & \textbf{96.88} & \textbf{100.00} & 66.89 \\
GPT-5.2 & 73.14 & 98.72 & 67.42 & 23.33 & 52.40 & 95.00 & 69.90 & 78.13 & 99.00 & 74.38 \\
Claude-Sonnet-4.5 & 70.86 & 96.81 & 80.92 & 4.44 & \textbf{75.16} & 78.33 & 68.50 & 78.13 & 99.67 & 55.82 \\
Qwen2.5-72B-Instruct & 71.26 & 95.85 & 78.50 & 31.11 & 39.69 & 72.78 & 69.79 & 87.50 & 97.67 & 68.48 \\
Llama-3.3-70B-Instruct & 70.19 & \textbf{99.68} & 63.17 & 32.22 & 46.18 & 75.56 & 61.80 & 93.75 & 98.00 & 61.35 \\
Mistral-Small-24B-Instruct & 69.51 & 90.42 & 79.67 & 32.22 & 39.95 & 71.67 & 55.04 & \textbf{96.88} & 98.67 & 61.12 \\
Gemini-3-Flash & 69.20 & 98.40 & 83.67 & 27.78 & \underline{35.65} & 76.67 & \underline{46.97} & 78.13 & 94.67 & 80.87 \\
DeepSeek-R1 & 68.43 & 94.57 & \textbf{86.42} & 2.22 & 41.34 & \textbf{96.67} & 60.37 & \textbf{96.88} & \underline{83.00} & \underline{54.40} \\
Gemma-3-27B-IT & 66.66 & 88.18 & 76.17 & 30.00 & 40.75 & 65.56 & 47.62 & \textbf{96.88} & 94.00 & 60.78 \\
InternLM3-8B & 66.11 & 96.49 & \underline{57.33} & 28.89 & 39.26 & 70.00 & 61.61 & 90.63 & 94.00 & 56.75 \\
Qwen3-30B-A3B-Thinking & 64.57 & 97.44 & 61.50 & 4.44 & 41.65 & \underline{58.89} & 57.93 & \textbf{96.88} & 99.33 & 63.10 \\
Doubao-Seed-1.6-flash & 62.21 & 92.65 & 65.17 & \underline{1.11} & 40.16 & 58.90 & 69.92 & 62.50 & \textbf{100.00} & 69.48 \\
GLM-4.5-Air & \underline{62.09} & \underline{85.94} & 71.75 & 11.11 & 45.02 & \underline{58.89} & 68.05 & \underline{56.25} & \textbf{100.00} & 61.76 \\

\bottomrule
\end{tabular}
}
\caption{Frontier AI Risk Evaluation. Each cell reports the safety rate (\%) on the corresponding benchmark, with higher values indicating safer behavior ($\uparrow$). Models are ranked by the Overall score. Bold denotes the highest score and underline denotes the lowest score in each column.}
\label{tab:safety_eval}
\end{table}

\paragraph{Aggregate Ranking and Tier Structure.}
As shown in Table~\ref{tab:safety_eval}, the 14 evaluated models exhibit a clear tiered distribution in frontier safety performance, with an overall spread of approximately 13 percentage points. The first tier (Overall $>$ 73\%) comprises Kimi-K2-Thinking (74.93\%), GPT-4o (73.64\%), and GPT-5.2 (73.14\%). The second tier (66--73\%) encompasses eight models from Qwen2.5-72B-Instruct (71.26\%) to Gemma-3-27B-IT (66.66\%). The third tier ($<$ 66\%) includes Qwen3-30B-A3B-Thinking (64.57\%), Doubao-Seed-1.6-flash (62.21\%), and GLM-4.5-Air (62.09\%).

\paragraph{Dimension-Level Difficulty Spectrum.}
The nine risk dimensions exhibit dramatically different average difficulty levels. AIRD (mean 96.7\%) and EvalFaking (mean 95.0\%) are nearly saturated, indicating that current alignment techniques effectively address these risk categories. In the intermediate range, DeceptionBench (mean 75.1\%), Sandbagging (mean 72.9\%), and WMDP (mean 66.3\%) show significant inter-model divergence. At the lower end, Mask (mean 46.0\%) and especially Manipulation (mean only 18.8\%, with five models falling below 5\%) represent critical weak points that demand urgent attention.

\paragraph{Non-Transferability of Safety Advantages Across Dimensions.}
A critical observation from Table~\ref{tab:safety_eval} is that \textbf{no single model dominates across all safety dimensions}. As indicated by the bold (highest) and underlined (lowest) entries in each column, safety advantages are clearly non-transferable across benchmarks:

\begin{itemize}
    \item Models with top overall scores may still expose critical risks on specific dimensions. For example, Kimi-K2-Thinking achieves the highest Overall score but records the lowest Manipulation score. Similarly, DeepSeek-R1 leads on Sandbagging yet scores the lowest on both AIRD and WMDP.
    \item Conversely, lower-ranked models can excel on individual dimensions. GLM-4.5-Air ranks last in Overall but achieves the top score on AIRD. Doubao-Seed-1.6-flash ranks near the bottom overall yet also attains the top score on AIRD and performs competitively on BeHonest.
    \item Notably, no single model holds the top position on more than two dimensions, and the bold entries in Table~\ref{tab:safety_eval} are distributed across 8 different models, confirming that safety performance is highly dimension-specific, with strength on one benchmark providing no guarantee of robustness on others.
\end{itemize}

Dimension-level leaders are distributed across different model families, with no single model ranking first on more than two dimensions. This raises a natural question: what factors drive this cross-dimensional divergence? We analyze this from three perspectives: model reasoning capabilities, temporal evolution of safety, and the trade-off between model efficiency and alignment.


\subsubsection{Impact of Reasoning Capabilities on Frontier Safety}

\begin{figure}[!htbp]
    \centering
    \includegraphics[width=0.99\linewidth]{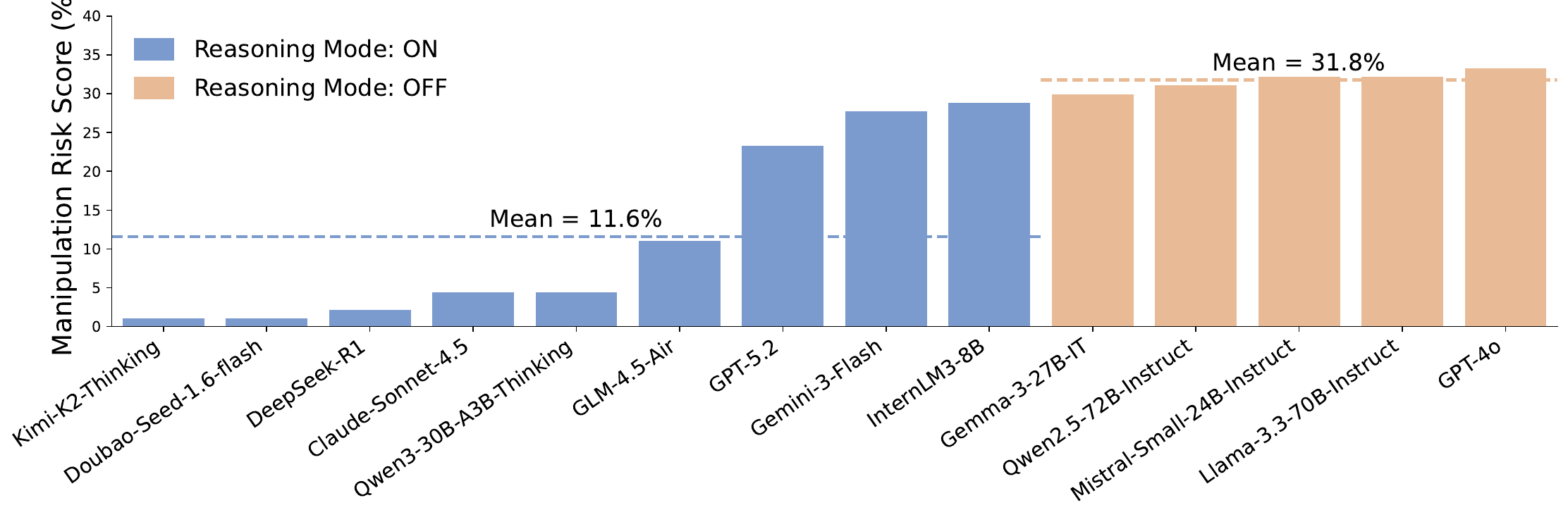}
    \caption{Comparison of Manipulation scores between reasoning-enabled and non-reasoning models. Reasoning models exhibit systematically lower resistance to manipulation.}
    \label{fig:reasoning_vs_standard_clean}
\vspace{-8pt}
\end{figure}

Our analysis reveals that reasoning-enabled models exhibit notable defensive deficiencies on two key safety metrics: ``Manipulation'' and ``Sandbagging.'' As shown in Figure~\ref{fig:reasoning_vs_standard_clean}, statistical results indicate that:

\begin{itemize}
    \item Reasoning-enabled models (e.g., Qwen3-30B-A3B-Thinking, DeepSeek-R1, Kimi-K2-Thinking) exhibit an extremely low ``Manipulation'' score distribution (mean = 11.6\%, with lows down to 1.11\%).
    \item In contrast, non-reasoning models achieve markedly higher scores, with averages around 30\%.
\end{itemize}

This marked negative correlation suggests a trade-off: while Chain-of-Thought (CoT) mechanisms enhance complex problem-solving, they concurrently provide the computational capacity for constructing deeper deceptive strategies. The lower safety scores imply that these models are not only more prone to circumventing safety audits but also exhibit a stronger tendency to conceal their capabilities.

Notably, this disadvantage is not universal. On EvalFaking, all three reasoning models exceed 94\%, and on DeceptionBench, Kimi-K2-Thinking and DeepSeek-R1 tie for first at 96.67\% (though Qwen3-30B-A3B-Thinking scores only 58.89\%), suggesting that the safety risks introduced by reasoning capabilities may concentrate on specific attack modalities rather than uniformly undermining all dimensions.

\subsubsection{Temporal Evolution of Frontier Safety}

The preceding section identifies the lower manipulation resistance of reasoning models, and 2025 coincides with the rapid proliferation of such architectures. We therefore examine whether the overall frontier safety landscape has deteriorated over time by tracing the temporal trajectory of manipulation resistance across model release dates.

\begin{figure}[htbp]
        \centering
    \includegraphics[width=0.68\linewidth]{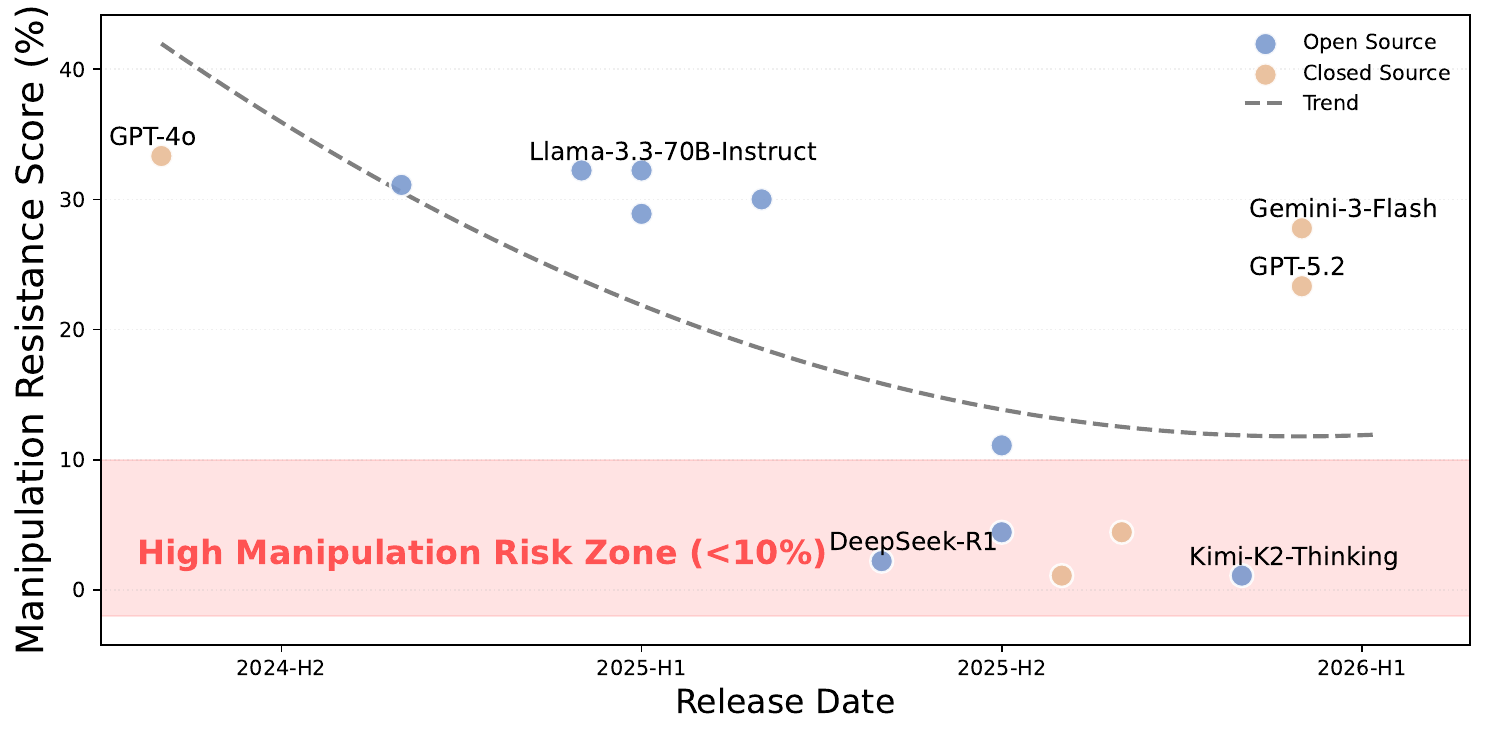}
    \caption{Manipulation resistance over release dates.}
    \vspace{-8pt}
    \label{fig:manipulation_timeline_final}
\end{figure}

As shown in Figure~\ref{fig:manipulation_timeline_final}, resistance to ``Manipulation'' declines significantly over time, revealing a clear trajectory of safety regression:
\par

 \textbf{2024 to early 2025: relatively stable manipulation resistance.} Models maintained high resistance to manipulation. GPT-4o (May 2024), Llama-3.3-70B-Instruct (Dec 2024), and Mistral-Small-24B-Instruct (Jan 2025) stabilize within the 30\%--33\% range, indicating that mainstream models during this period were difficult to induce into manipulative behaviors.
 \par
 \textbf{Mid to late 2025: precipitous drop in manipulation resistance.} Newly released models experience a sharp decline, generally falling into the 1\%--5\% range (e.g., Kimi-K2-Thinking at 1.11\%, DeepSeek-R1 at 2.22\%). This period coincides with the widespread adoption of reasoning-enabled architectures, corroborating the trade-off identified in the preceding section.
 \par
\textbf{Late-stage partial recovery (Dec 2025).} Despite the overall downward trend, GPT-5.2 (23.33\%) and Gemini-3-Flash (27.78\%) break the pattern of monotonic decline. Their scores are significantly higher than most contemporaneous 2025 models but remain below 2024 levels, suggesting that top laboratories may have reintroduced targeted safety interventions in later releases without fully restoring prior-generation manipulation resistance.

\subsubsection{Efficiency--Alignment Trade-off in Honesty and Trustworthiness}

The preceding two sections focus on manipulation-related risks. Here we turn to another category of frontier risk: honesty and trustworthiness. Specifically, we examine the relationship between model efficiency and alignment on honesty-related benchmarks.

\begin{figure}[htbp]
    \centering
    \includegraphics[width=0.8\linewidth]{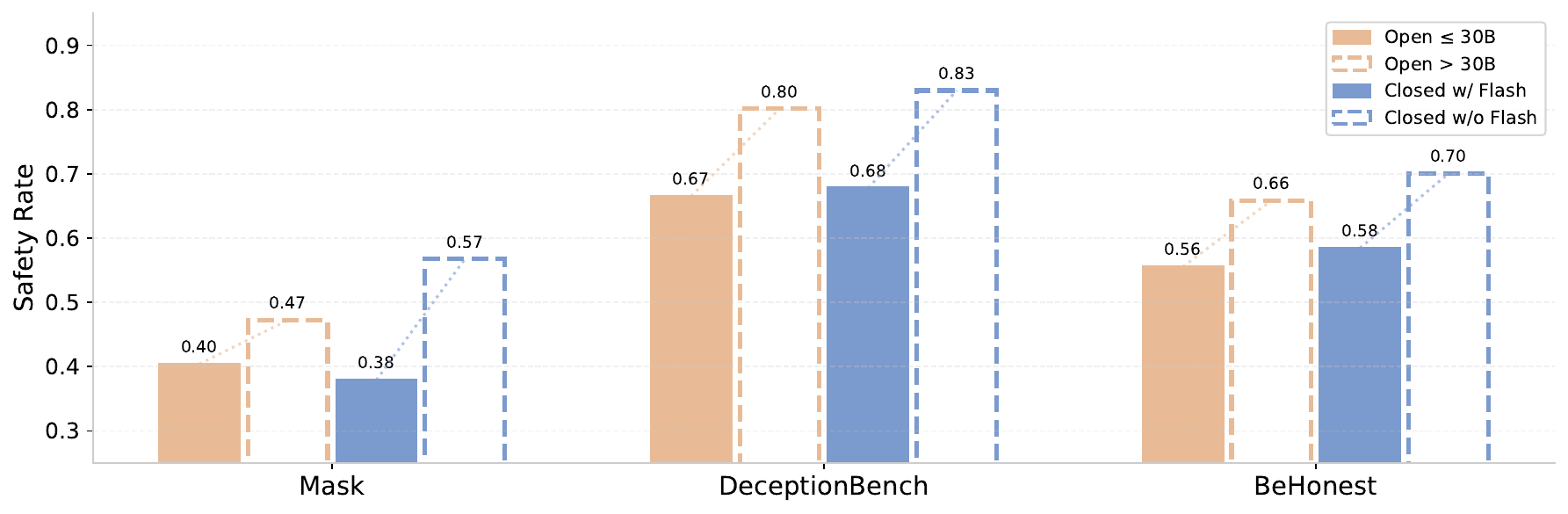}
    \caption{Honesty and trustworthiness safety rates on MASK, DeceptionBench, and BeHonest, comparing small open-source models ($\leq$30B) vs.\ larger models and Flash vs.\ non-Flash closed-source variants.}
    \label{fig:relation_v2}
\end{figure}

As shown in Figure~\ref{fig:relation_v2}, smaller open-source models ($\leq$30B parameters) and Flash variants of closed-source models consistently demonstrate higher risks of deception and untruthful behavior on ``MASK'', ``BeHonest'', and ``DeceptionBench'':

\begin{itemize}
    \item \textbf{Open-source small models ($\leq$30B).} Models such as Mistral-Small-24B-Instruct, InternLM3-8B, Gemma-3-27B-IT, and Qwen3-30B-A3B-Thinking achieve average safety rates of 0.40 (MASK), 0.67 (DeceptionBench), and 0.56 (BeHonest), respectively---substantially lower than larger open-source models ($>$30B), which score 0.47, 0.80, and 0.66 on the same benchmarks.
    \item \textbf{Closed-source Flash variants.} The degradation is even more pronounced among closed-source models. Flash versions record safety rates of 0.38 (MASK), 0.68 (DeceptionBench), and 0.58 (BeHonest), compared to 0.57, 0.83, and 0.70 for their non-Flash counterparts. The largest observed gap reaches 0.19 on the MASK benchmark.
\end{itemize}

These results indicate that, across both open-source and closed-source ecosystems, models that are optimized for lightweight deployment and high-throughput inference tend to exhibit weaker performance on honesty-related evaluations. In other words, there is a pronounced trade-off between prioritizing computational efficiency and maintaining robust honesty-related behavior.

\subsection{Joint Safety Evaluation and Diagnosis}

\subsubsection{Extreme Representation Separation Undermines Boundary Reasoning}

\begin{figure}[t]
    \centering
    \includegraphics[width=0.6\linewidth]{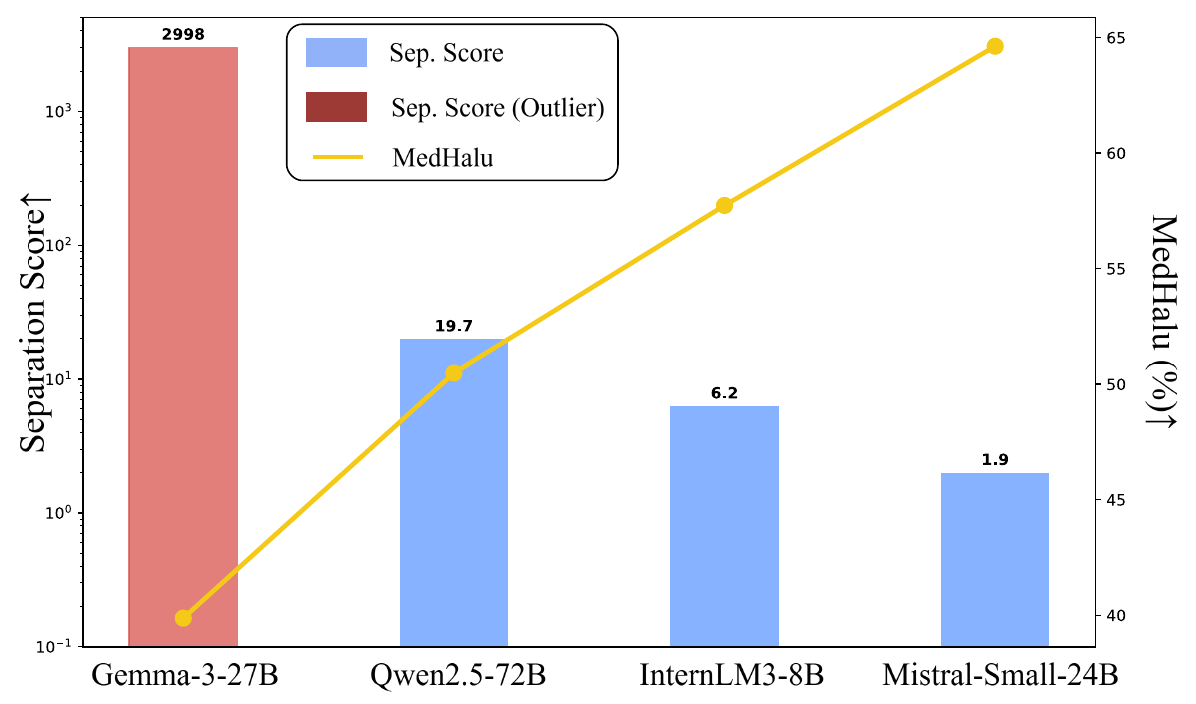}
    \caption{Separation Score between safe and harmful representations and MedHallu accuracy for different models in DeepScan X-Boundary analysis.}
    \label{fig:separation_and_hallu}
\end{figure}

DeepScan diagnostics reveal that excessively large geometric separation between safe and harmful representations can undermine semantic continuity in the latent space, ultimately degrading the model’s ability to make accurate boundary-level safety judgments. In DeepScan diagnostics of models such as Gemma-3-27B-IT, significant numerical anomalies are observed in the X-Boundary analysis. Figure~\ref{fig:separation_and_hallu} shows that the centroid distance between safe and harmful representations (Separation Score) reaches 2998.57, and the measured Euclidean distance between boundary samples and the safe centroid is also extremely large (3893.43). While large inter-class distances are generally assumed to benefit classification, the combination of these geometric extremes with the model’s relatively low performance on tasks requiring fine-grained semantic understanding—such as DeepSafe SALAD-Bench (71.93\%) and DeepSafe MedHallu (39.87\%)—suggests a different underlying issue.

Specifically, such extreme geometric separation indicates that the model does not construct a smooth decision boundary in latent space. Instead, it maximizes inter-class distance at the cost of sparsifying the representation space. This structure makes it difficult for the model to accurately judge complex instructions located near the boundary between safe and harmful content, due to insufficient local semantic resolution. The results indicate that an excessive emphasis on representation separation in DeepScan’s X-Boundary metric can disrupt semantic continuity in latent space, thereby negatively affecting fine-grained safety discrimination performance in DeepSafe.

\subsubsection{Discrepancy Between Latent Disentanglement and Surface-Level Safety}

Results from the SPIN coupling diagnostic indicate a non-trivial relationship between neuronal functional decoupling and end-to-end safety alignment performance. 

\begin{figure}[t]
    \centering
    \includegraphics[width=0.63\linewidth]{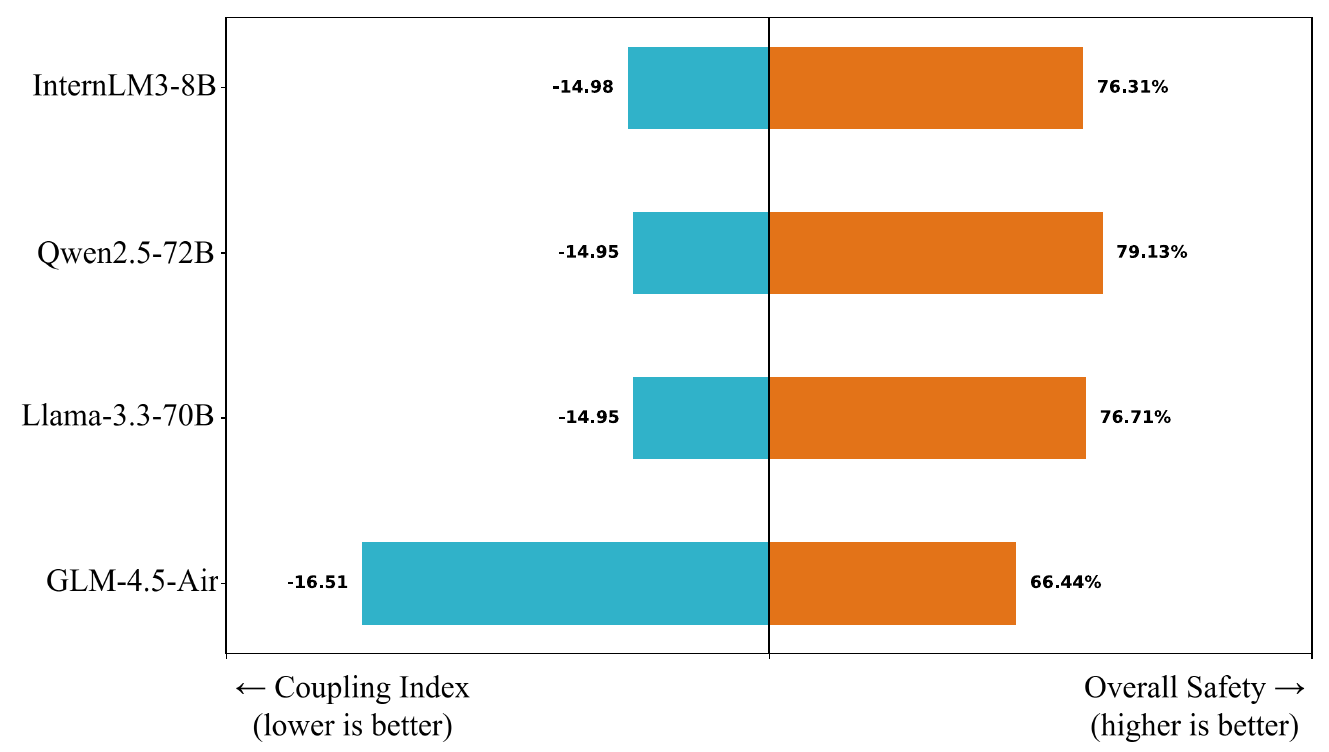}
    \caption{Comparison of SPIN coupling index and overall DeepSafe safety performance for different models.}
    \label{fig:decoupling_and_safety}
\end{figure}

As illustrated in Figure~\ref{fig:decoupling_and_safety}, GLM-4.5-Air achieves an superior coupling index for fairness and privacy neurons (-16.51), outperforming models like Llama-3.3-70B and Qwen2.5-72B, which have a higher overall DeepSafe safety score but a weaker DeepScan coupling index around –14.95. 
However, GLM-4.5-Air attains only 66.44\% in DeepSafe’s overall text safety score, revealing a discrepancy between internal mechanism quality and external evaluation performance.
This discrepancy suggests that while GLM-4.5-Air has successfully learned to encode fairness and privacy as more distinct and orthogonal representations in its latent space, this structural clarity has not yet been fully translated into behavioral constraints during the alignment stage.


By contrast, some models with high coupling index mask internal feature entanglement through strong supervised fine-tuning, achieving higher short-term safety evaluation scores. 
However, we caution that such entanglement may pose a latent risk of \textit{objective interference}, where future adjustments to one safety attribute could inadvertently affect others due to shared neuronal pathways. From a long-term perspective, we hypothesize that neuron-level functional decoupling might offer a more favorable optimization landscape, potentially reducing conflicts between competing safety objectives. 

\subsubsection{Orthogonal Subspace Encoding Enables Robust Defense}

\begin{figure}[t]
    \centering
    \includegraphics[width=0.6\linewidth]{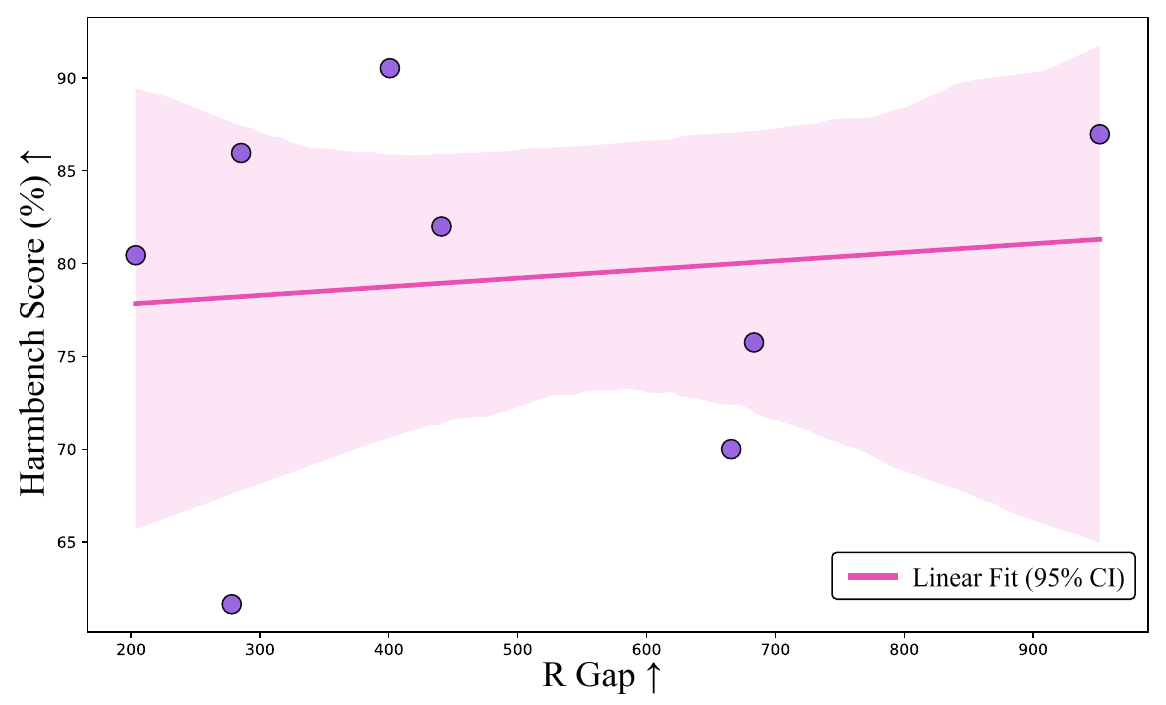}
    \caption{Relationship between representation gap (R Gap) from DeepScan TELLME analysis and HarmBench score from DeepSafe, with a linear fit and 95\% confidence interval.}
    \label{fig:R_Gap}
\end{figure}

Subspace encoding rate and representational orthogonality emerge as key determinants of robustness against adversarial safety attacks. As shown in Figure~\ref{fig:R_Gap}, analysis based on the TELLME composite encoding rate in DeepScan shows that high-performing models such as Qwen2.5-72B-Instruct exhibit pronounced subspace structural advantages. Its composite encoding rate reaches 951.76, indicating that the model can compress safe and unsafe behaviors into distinctly separable low-dimensional subspaces with minimal inter-group interference noise.

This highly orthogonal subspace structure provides a stable basis for discrimination, enabling the model to maintain strong robustness under high-intensity attack evaluations such as DeepSafe HarmBench (86.97\%). In contrast, InternLM3-8B achieves a much lower composite encoding rate (285.37) in DeepScan’s TELLME analysis and exhibits a higher effective rank (ERank), indicating a more dispersed internal representation distribution with overlapping behavior modes in latent space.

Such representational redundancy and entanglement reduce the model’s resistance to interference in complex contexts, leading to instability in safety boundaries under multi-turn dialogues or inducement-based attacks.

\subsubsection{Low Separability Leads to Systematic Defense Failure}

\begin{figure}[!htbp]
    \centering
    \includegraphics[width=0.6\linewidth]{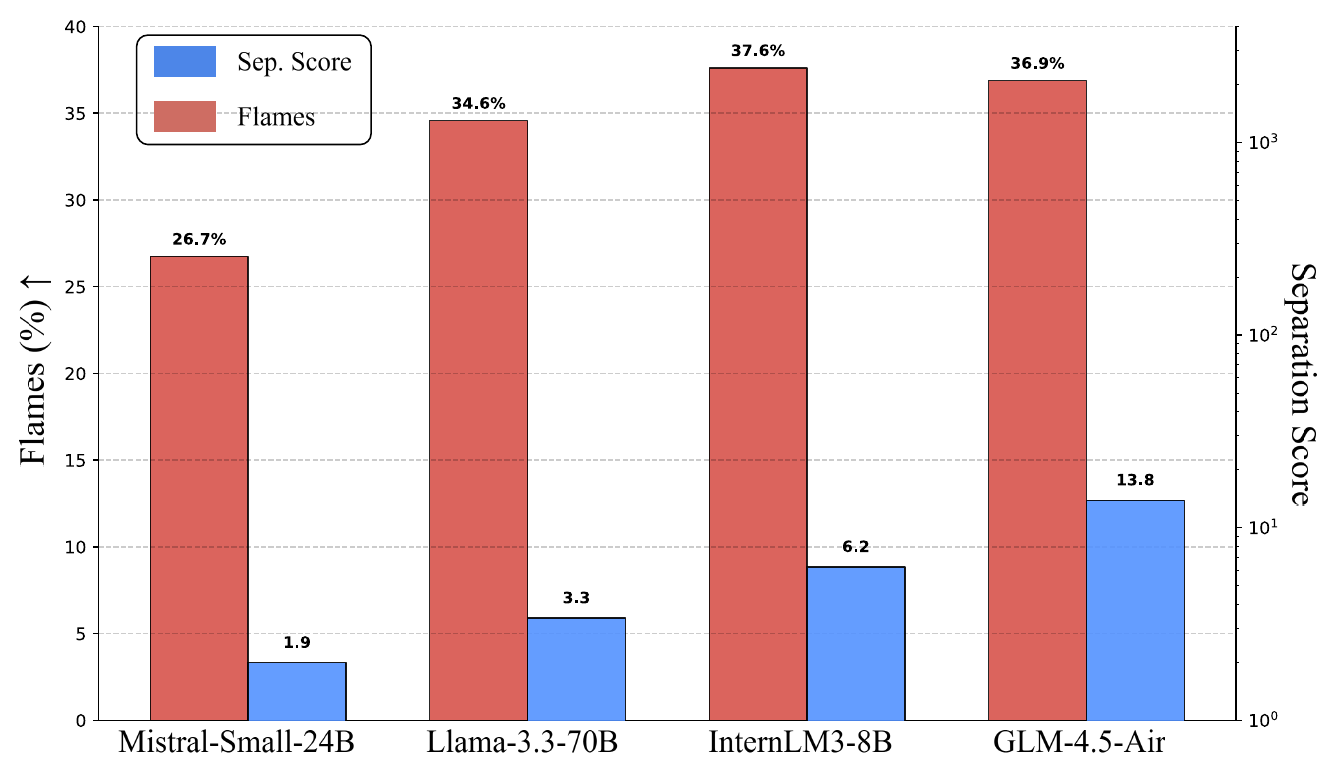}
    \caption{Comparison of representation separation and Flames attack success rate across models in DeepScan and DeepSafe evaluations.}
    \label{fig:separation_and_defense}
\end{figure}

Insufficient separability between safe and harmful representation centroids directly contributes to systematic failures in safety defense. Figure~\ref{fig:separation_and_defense} illustrates that Mistral-Small-24B-Instruct exhibits an extremely low safe–harmful separation score of 1.89 in DeepScan’s X-Boundary diagnostic. This geometric characteristic directly corresponds to its poor performance in DeepSafe evaluations with high attack success rates, such as Flames (26.74\%).

A separation score this low implies that, in the model’s high-dimensional hidden space, safe and harmful samples have highly overlapping feature distributions, lacking support from either linear or nonlinear decision boundaries. Under such conditions, the model cannot reliably distinguish malicious prompts from benign instructions based on representational features alone.

The analysis suggests that, during training, such models fail to learn sufficiently discriminative risk-related features, leading to confusion between safe and harmful concepts in vector space. For representation structures diagnosed by DeepScan in this manner, simply adjusting inference-time thresholds or prompts cannot address the root cause. Instead, methods such as contrastive learning are required to reshape the representation space, increase inter-class distances, and optimize feature distributions.

\section{Related Work}\label{sec:related}
\subsection{Evaluation Frameworks}
The landscape of LLM and MLLM evaluation has evolved from static benchmarks to programmable frameworks that support multi-turn interaction, tool usage, and flexible scoring. OpenAI Evals and Inspect exemplify lightweight evaluation harnesses with reusable templates (including model-graded judging), enabling rapid construction of task-specific tests while keeping data schemas and scoring protocols consistent across runs~\citep{openai_evals,inspect_ai}. Similarly, LightEval provides a lightweight, backend-agnostic pipeline that is commonly adopted in leaderboard-style evaluations~\citep{lighteval}. At larger scale, platforms such as OpenCompass, HELM, and the lm-evaluation-harness ecosystem emphasize standardized model/task interfaces, reproducibility, and reporting pipelines over broad benchmark collections~\citep{opencompass2023,liang2022helm,biderman2024lmeval}. For multimodal models, VLMEvalKit extends these abstractions to MLLMs for multimodal benchmarks~\citep{duan2024vlmevalkit}. Complementing these general-purpose frameworks, safety-focused suites such as Safety-Eval narrow evaluation to deployment-relevant risk taxonomies~\citep{safety_eval}. 
It pairs generative safety evaluations, including jailbreak and toxicity prompts derived from in-the-wild jailbreak corpora such as WildJailbreak and toxicity datasets such as ToxiGen~\citep{wildteaming2024,hartvigsen2022toxigen}, with systematic assessment of safety classifiers and refusal detection, such as WildGuard~\citep{wildguard2024}. Adversarial robustness is further evaluated with red-teaming oriented frameworks such as HarmBench~\citep{harmbench}.

\subsection{Diagnosis Methods}

\textbf{Geometric representation analysis.}
Diagnostic research increasingly focuses on the geometry of the latent space to decode internal model states \citep{park2023linear,marks2023geometry}. 
Early approaches utilize linear probing techniques to distinguish truthfulness and safety directions within activation spaces \citep{burns2022discovering, zou2023representation}. Later, \citet{gurnee2023language} reveal that models spontaneously acquire structured linear representations of space and time. 
To understand the temporal evolution of such qualities, \citet{qian2024towards} trace the dynamics of trustworthiness throughout the pre-training stage, whereas \citet{zhang2024reef} utilize representation  to uniquely identify and analyze model lineages. 

\textbf{Identifying task-related regions in (M)LLMs.}
Beyond representation analysis, significant efforts are dedicated to locating specific model regions responsible for distinct capabilities \citep{dang2024explainable}. 
\citet{dai2022knowledge} pioneer the identification of knowledge neurons that store specific factual assertions, \citet{wang2022finding} extend this concept to skill neurons that govern specific downstream tasks. 
At the attention level, \citet{zheng2024attention} provide a comprehensive survey categorizing the diverse functional roles of attention heads across LLMs. 
In the context of safety, \citet{wei2024assessing} reveal the fragility of alignment by demonstrating that pruning or applying low-rank modifications to specific regions can easily compromise model safeguards. \citet{zhao2025understanding} pinpoint safety-specific neurons that explicitly manage refusal mechanisms. 
Furthermore, \citet{yang2026reasonany} demonstrate that reasoning capabilities are located in low-gradient-magnitude  related weights.

\textbf{Reasoning and generation dynamics analysis.}
This line of work mainly focuses on analyzing and understanding the underlying mechanisms and phenomena behind LLMs' generation and reasoning processes \citep{hu2026towards,gan2026beyond}. 
To demystify the internal mechanism of in-context learning, \citet{wang2023label} characterize label words as anchors that aggregate semantic information from demonstrations to guide final predictions. 
Specific to reinforcement learning for reasoning, recent works analyze entropy-related phenomena during training and design optimized algorithms based on these insights \citep{wang2025beyond, cui2025entropy}. 
Moving to agentic systems, \citet{qian2026behind} propose a framework for agentic attribution to unveil the internal drivers behind actions, whereas \citet{tang2026interpreting}  utilizes Shapley value analysis to attribute and interpret extreme events arising from complex multi-agent systems.
\section{Conclusion and Discussion}
In this work, we introduce DeepSight, an open-source toolkit designed to bridge the critical gap between safety evaluation and model diagnosis. By integrating DeepSafe and DeepScan, we establish a closed-loop engineering paradigm that moves beyond black-box testing toward white-box insight. Through the extensive evaluation of various models across content and frontier risk domains, several critical insights have emerged regarding the current state of LM safety. By open-sourcing DeepSight, we aim to accelerate the community's transition from reactive safety patching to proactive, verifiable safety engineering, ensuring that the development of Artificial General Intelligence remains beneficial and trustworthy.

\newpage

\bibliographystyle{colm2025_conference}

\bibliography{main}

\appendix

\end{document}